\documentclass{article} % For LaTeX2e
\usepackage{iclr2027_conference,times}
\iclrfinalcopy

\usepackage{amsmath,amsfonts,bm}

\def\eqref#1{equation~\ref{#1}}
\def\1{\bm{1}}

\DeclareMathAlphabet{\mathsfit}{\encodingdefault}{\sfdefault}{m}{sl}
\SetMathAlphabet{\mathsfit}{bold}{\encodingdefault}{\sfdefault}{bx}{n}

\usepackage{amsthm}
\usepackage{enumitem}
\usepackage{hyperref}
\usepackage{url}
\usepackage{notation}
\usepackage{researchpack}
\usepackage{booktabs}
\usepackage{multirow}
\usepackage[table]{xcolor}
\usepackage{graphicx}
\usepackage{tabularx}
\usepackage{algorithm}
\usepackage{wrapfig}
\usepackage{algpseudocode}
\definecolor{lightgreen}{RGB}{223,240,216}
\definecolor{posgreen}{HTML}{1C7935}
\usepackage[capitalize,noabbrev]{cleveref}
\theoremstyle{plain}
\newtheorem{theorem}{Theorem}[section]
\newtheorem{proposition}[theorem]{Proposition}

\theoremstyle{definition}
\newtheorem{definition}[theorem]{Definition}

\title{Learning to Couple: Context-Adaptive Joint \\ Denoising for Diffusion Language Models}
\author{
\parbox{\textwidth}{
\vspace{1em}
Zilei Shao\textsuperscript{1}\thanks{Equal contribution.}\quad
Ian Li\textsuperscript{2}\footnotemark[1]\quad
Heng Zhao\textsuperscript{3}\quad
Kewei Lian\textsuperscript{4}\\[0.5em]
Guy Van den Broeck\textsuperscript{1}\quad
Zhe Zeng\textsuperscript{3}\quad
Anji Liu\textsuperscript{4}\thanks{Corresponding author.}\\[0.8em]
\normalfont
\textsuperscript{1}University of California, Los Angeles\quad
\textsuperscript{2}University of California, San Diego\\
\textsuperscript{3}University of Virginia\quad
\textsuperscript{4}National University of Singapore
}
\vspace{-1em}
}

\begin{document}

\maketitle
\lhead{Preprint}

\begin{abstract}
Diffusion language models accelerate generation by predicting multiple tokens in parallel, but factorized denoisers treat simultaneous predictions as conditionally independent. Decoding heuristics reduce the resulting errors by controlling which tokens are committed, but leave the underlying distribution factorized. Joint modeling addresses the source of the problem, yet existing methods attach separately learned layers to frozen denoisers, leaving the backbone unadapted to the joint objective and the coupling parameters fixed across contexts. We propose Codpula, a context-parameterized denoiser combining a Transformer with a tractable probabilistic inference layer. Given a partially masked input, the Transformer produces token potentials that shape per-position predictions and compact gates that modulate the layer's coupling parameters, yielding a context-dependent joint distribution while retaining exact conditional inference. We jointly optimize both components under a conditional denoising objective using a dual-flow multiplicative update that preserves valid probabilistic parameters. At inference, Codpula supports standard heuristics and uses conditional queries to adapt commitments to dependencies among proposed tokens without extra Transformer evaluations within a step. Codpula sets new state-of-the-art results across a broad suite of reasoning benchmarks while advancing the accuracy–latency Pareto frontier in mathematical reasoning and code generation.
\end{abstract}

\section{Introduction}

Masked diffusion language models can accelerate generation by predicting multiple tokens in parallel, reducing the number of denoising steps. However, standard denoisers factorize their predictions across positions \citep{austin2021structured,sahoo2024simple,nie2025large}. Even under a shared context, these predictions remain conditionally independent, preventing the model from expressing dependencies among tokens predicted together. These missing dependencies become increasingly consequential as more tokens are committed at once \citep{liu2025discrete}. A parallel denoiser must therefore predict both \emph{what} each token could be and \emph{how} its predictions depend on one another.

Many approaches address the factorization barrier during decoding by varying commitments, prioritizing confident predictions, or revising earlier tokens \citep{arriola2025block,wu2025fast,ben-hamu2025accelerated,zhao2024informed,israel2025accelerating}. These methods improve the quality–computation trade-off while retaining a factorized denoiser. Joint modeling instead captures dependencies among simultaneous predictions \citep{li2026breaking,kim2026tensortrainjointmodelingfewstep}. Existing models, however, typically attach a separately trained joint layer to a frozen backbone. This layer produces context-dependent probabilities, but its dependency parameters remain fixed across inputs and the backbone is not optimized through the joint distribution.

\begin{figure}
    \centering
    \includegraphics[width=\linewidth]{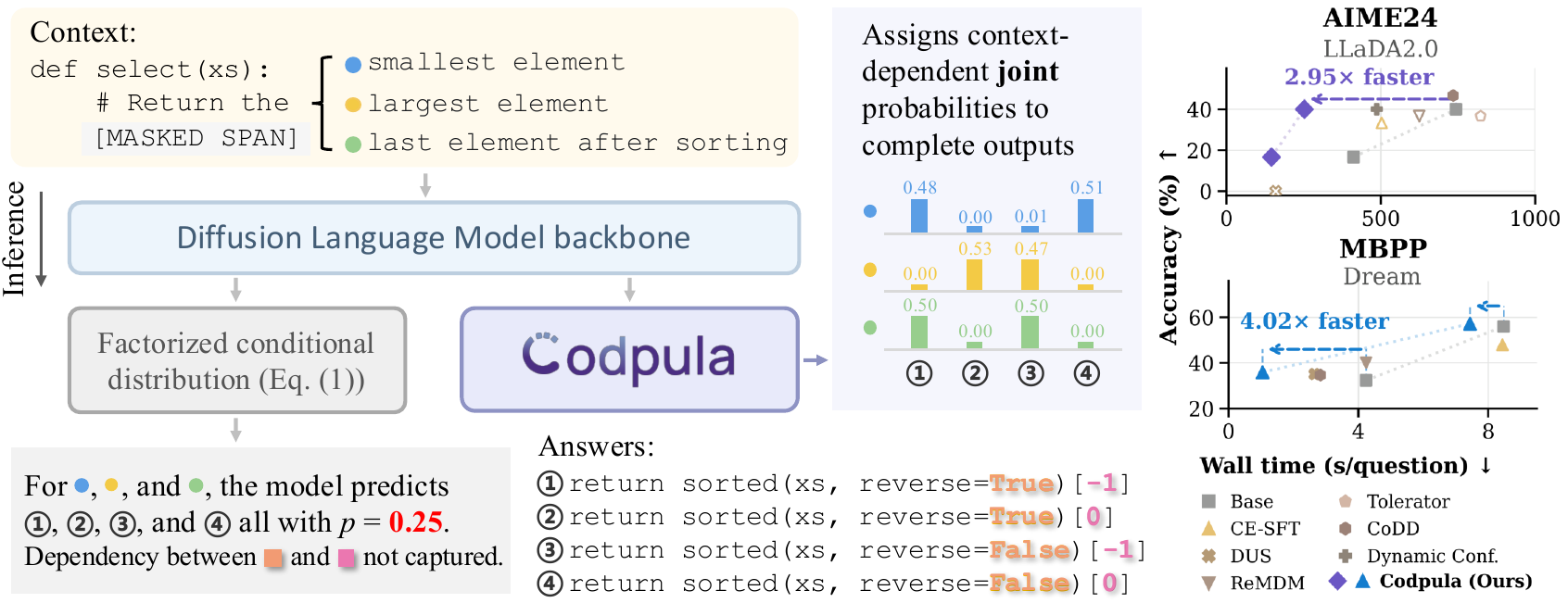}
    \vspace{-2.0em}
    \caption{\textbf{Factorized versus context-dependent joint prediction.} \textbf{Left:} Given a masked return statement, a factorized denoiser predicts the sorting flag and index independently, assigning equal probability to all four completions regardless of the comment. \textbf{Middle:} Codpula assigns context-dependent joint probabilities, placing its mass on the completions consistent with each comment. \textbf{Right:} Codpula improves accuracy--latency trade-offs, achieving $2.95\times$ and $4.02\times$ speedups at matched accuracy on AIME24 (LLaDA-2.0; \citet{bie2025llada20}) and MBPP (Dream; \citet{ye2025dream}).}
    \label{fig:fig1}
    % \vspace{-0.4em}
\end{figure}

We propose Codpula, a jointly learned, context-parameterized denoiser that adapts both token predictions and their coupling to the current context. From the backbone representation, Codpula derives token scores that adjust per-position predictions and a compact set of gates that reweight shared coupling parameters. The resulting joint distribution can favor different combinations of token values depending on the context, as illustrated in \cref{fig:fig1} (left): Codpula assigns high probability to different combinations of sort direction and index for different requested operations. Training through this distribution teaches the backbone to produce token preferences and coupling signals that together explain the clean completion. To support this training, we derive a dual-flow update that accommodates the signed learning statistics introduced by conditional normalization while preserving valid probabilistic parameters.

Beyond modeling which token values belong together, the learned joint distribution informs which proposals to commit together. Codpula supports the token-selection heuristics used by existing block and full-sequence decoders and provides conditional queries for dependence-adaptive decoding. These queries measure how each proposal's probability changes when the other candidates are assigned their proposed values. The decoder defers sensitive proposals, adapting the number of commitments to their dependencies. All queries reuse the token potentials and gates from the same Transformer forward pass, requiring no additional backbone evaluation within the step.

We evaluate Codpula on mathematical reasoning and code generation across four modern diffusion language models against a comprehensive suite of state-of-the-art baselines. The results show improved accuracy across a range of parallel decoding budgets and improved accuracy-latency trade-offs after accounting for inference overhead. \cref{fig:fig1} (Right) highlights that Codpula achieves comparable accuracy with $2.95\times$ faster generation on AIME24 using LLaDA-2.0
and $4.02\times$ faster generation on MBPP using Dream.

\section{Preliminary}
\subsection{Diffusion Language Models}
Diffusion language models (dLLMs) learn to generate text from partially masked input by iteratively reconstructing masked portions of an input sequence \citep{austin2021structured, sahoo2024simple, shi2024simplified}. Let $\x_0=(x_0^0,\ldots,x_0^L)\in\mathcal{V}^{L}$ denote a clean sequence, where $\mathcal V$ is a vocabulary. At timestep $t$, each token is independently masked with probability $\alpha_t$, yielding a partially masked sequence $\x_t = (x_t^0, \dots, x_t^L)$. Given the partially masked sequence $\x_t$, a neural denoiser parameterized by $\vparams$ predicts the original clean sequence $\x_0$ using a factorized conditional distribution
\begin{equation}
    p_{\vparams}(\x_0|\x_t)
    = \prod_{i=1}^{L}p_{\vparams}(x_0^i|\x_t).
    \label{eq:bg-factorized}
\end{equation}
Under this factorization, the denoising objective evaluates reconstruction only at masked positions, leaving the observed tokens unchanged. To construct a training example, we sample a clean sequence $\x_0\sim p_{\mathrm{data}}$ and a timestep $t$, then obtain $\x_t$ by independently masking each token in $\x_0$ with probability $\alpha_t \in (0,1]$. The model is trained by maximizing the weighted denoising objective
\begin{equation}
    \calL(\vparams)
    = \Ex_{t,\x_0,\x_t}\!\left[
        w(t)\log p_{\vparams}(\x_0|\x_t)
      \right],
    \label{eq:bg-diffusion-objective}
\end{equation}
\noindent where $\x_0\sim p_{\mathrm{data}}$, and $w(t)$ is a weighting function \citep{sahoo2024simple, shi2024simplified}. 

 At inference, generation starts from a fully masked sequence and progressively reveals tokens as $t$ decreases. With enough denoising steps, the model can reveal one token at a time, allowing every prediction to condition on all earlier ones. In this sequential limit, \cref{eq:bg-factorized} represents the desired target distribution since only one token is predicted at each step. In practice, however, fast generation requires revealing multiple tokens per denoising step. \cref{eq:bg-factorized} treats these tokens as conditionally independent, even though they may remain strongly dependent given the observed context. The model is thus asked to predict a joint completion using a product of token-wise distributions, which can produce inconsistent outputs \citep{xu2025energy}. Even perfectly learned token marginals cannot generally recover a dependent conditional joint distribution through their product, so increasing backbone capacity alone does not remove this representational constraint \citep{liu2025discrete,hayakawa2025distillation}. Revealing fewer tokens per step can reduce the resulting dependence error, but requires more backbone evaluations \citep{ben-hamu2025accelerated}. We refer to this tension between parallel generation and factorized prediction as the \textbf{factorization barrier} \citep{li2026breaking}.

\subsection{Parallel Denoising is Conditional Joint Prediction}
Two broad approaches address the factorization barrier. The first retains the factorized outputs and compensates at inference. Heuristic-based decoding ranks positions by confidence, entropy, or margin \citep{chang2022maskgit, ye2025dream, kim2025train}; adaptive parallel decoding adjusts the number of tokens committed per step \citep{israel2025accelerating}; remasking enables revision of previously generated tokens \citep{wang2025remasking}. However, tokenwise scores do not explicitly measure dependencies among jointly proposed tokens \citep{wu2025fast}, while iterative refinement can trade additional model evaluations for improved sample quality \citep{zhao2024informed,wang2025remasking}.

The second approach confronts the barrier directly by modeling the conditional joint distribution of tokens predicted in parallel. Coupled Discrete Diffusion (CoDD) \citep{li2026breaking} and tensor-train joint modeling \citep{kim2026tensortrainjointmodelingfewstep}\ use structured joint parameterizations to capture token dependencies while supporting efficient inference. Both approaches combine neural token potentials with a coupling model for simultaneous predictions. Their joint distributions can be expressed as
\begin{equation}
    p_{\vparams,\wparams}(\x_0 | \x_t)
    =
    \frac{1}{Z(\x_t)} \cdot
    p_{\wparams}(\x_0 | \x_t) \cdot
    \prod_{i=1}^{L}p_{\vparams}(x_0^i \given x_t).
    \label{eq:bg-codd-product}
\end{equation}

\section{From Static to Context-Dependent Coupling}
\label{sec:joint-learning}

\cref{eq:bg-codd-product} reveals a limitation of existing joint predictors. The dLLM’s tokenwise predictions $p_{\vparams}(\x_0^i\mid \x_t)$ adapt to context, whereas coupling parameters $\wparams$ are fixed across inputs, mask patterns, and timesteps. A single coupling model must therefore capture dependencies in every denoising context, despite substantial variation in masked-token dependencies with the observed tokens.

The remedy is clear: token predictions and their dependencies must both adapt to the denoising context. We therefore propose the following context-dependent joint denoiser:
\begin{equation}
    \params_t=f_{\vparams}(\x_t),
\qquad
p_{\vparams,\wparams}(\x_0 \mid \x_t) := p(\x_0;\params_t,\wparams),
    \label{eq:codd2}
\end{equation}
\noindent where $\params_t$ contains both the tokenwise prediction parameters and the context-dependent coupling parameters, while $\wparams$ denotes the parameters intrinsic to the probabilistic model.

There are many ways to parameterize $p(\x_0;\params_t,\wparams)$, but the distribution must remain efficiently normalizable when $\params_t$ changes with $\x_t$. To make this requirement explicit, we write
\begin{align}
p(\x_0;\params_t,\wparams)
=
\frac{\tilde{p} (\x_0;\params_t,\wparams)}
{Z(\params_t,\wparams)},
\qquad
Z(\params_t,\wparams)
=
\sum_{\x_0}
\tilde{p} (\x_0; \params_t,\wparams),
\label{eq:Z}
\end{align}
\noindent where $\tilde{p}$ is an unnormalized joint probability. Because $\params_t$ changes at every denoising step, $Z(\params_t,\wparams)$ must be recomputed for each $\x_t$ and generally requires summing over exponentially many sequences. To overcome this challenge, we parameterize $\tilde{p}$ using a probabilistic circuit (PC) \citep{ProbCirc20}, whose sum-product structure enables exact and efficient computation of $Z(\params_t,\wparams)$ for dynamically generated $\params_t$. We next review background on PCs that make this computation tractable.

\paragraph{Probabilistic Circuits.}A probabilistic circuit (PC) represents a joint distribution through a rooted directed acyclic graph of input, sum, and product nodes~\citep{ProbCirc20, poon2011sum}. Input nodes define univariate distributions over variables $X \in \X$. Sum and product nodes recursively combine these distributions into a joint distribution over $\X$. Formally, each node computes
\begin{equation}
    p_n(\x)=
    \begin{cases}
        g_n(\x),
            & n\text{ is an input node over }X \in \X,\\
        \prod_{c\in\ch(n)}p_c(\x),
            & n\text{ is a product node},\\
        \sum_{c\in\ch(n)}\wparams_{n,c} \cdot p_c(\x),
            & n\text{ is a sum node},
    \end{cases}
    \label{eq:bg-pc}
\end{equation}
\noindent where $\ch(n)$ denotes the children of $n$, $g_n$ is a univariate distribution over $X \in \X$, and the nonnegative sum node parameters $\{\wparams_{n,c}\}_{n,c}$ satisfy $\sum_{c\in\ch(n)}\wparams_{n,c}=1$ ($\forall \text{~sum~node~} n$). The key advantage of PCs is their ability to compute many probabilistic queries exactly and efficiently \citep{vergari2021compositional}. This tractability makes them particularly well suited to our setting: when $\tilde{p}(\x_0;\params_t,\wparams)$ is represented by a PC satisfying suitable structural requirements, the normalization constant ($Z$) in \cref{eq:Z} can be computed exactly in a single bottom-up pass.\footnote{Appendix~\ref{app:pc-normalization} describes the specific structural properties of PCs and details of this bottom-up pass.}

We next turn to training this hybrid denoiser. Applying the standard diffusion-ELBO derivation \citep{sahoo2024simple, shi2024simplified} in this setting yields the following generalized denoising objective (see Appendix~\ref{app:elbo-derivation}, which we maximize jointly over the neural and circuit parameters):
\begin{equation}
    \calL(\wparams,\vparams)
    =\Ex_{t,\x_0,\x_t}\!\left[
        w(t)\log p_{\vparams,\wparams}(\x_0|\x_t)
      \right],
    \label{eq:codd-objective}
\end{equation}
\noindent where $t\sim\mathcal U(0,1)$, $\x_0\sim p_{\mathrm{data}}$, $\x_t\sim q_t(\cdot|\x_0)$, and $w(t)\geq0$ is the diffusion weighting function. 

Although \cref{eq:codd-objective} is differentiable end to end, its parameter blocks require different optimization treatments. The neural parameters ($\vparams$) can be updated using standard gradient-based optimizers. The circuit parameters ($\wparams$), however, require expectation-maximization-style (EM-style) updates, whose constraint-preserving behavior relies on nonnegative log-parameter derivatives \citep{peharz2020einsum,liu2025rethinkingprobabilisticcircuitparameter}. While this condition holds for standalone PCs, the gradients induced by our objective can be signed, so the standard EM-style update does not apply directly. To make this issue precise, we first recall a general formulation that encompasses commonly used EM-style updates: 
\begin{equation}
    \wparams^{\mathrm{new}}_{n,c}
    \propto \wparams_{n,c}+\alpha \cdot T_{\vparams, \wparams} (n,c) \cdot \frac{\partial\calL(\wparams,\vparams)}{\partial \log \wparams_{n,c}},
    \label{eq:anemone}
\end{equation}
\noindent where $\alpha \geq 0$ is the base learning rate, and $T_{\vparams,\wparams}(n,c)$ is an optional per-parameter step-size multiplier. Varying the batch, $\alpha$, and $T_{\vparams,\wparams}$ recover full-batch EM, mini-batch EM, and Anemone \citep{liu2025rethinkingprobabilisticcircuitparameter} as special cases (Appendix~\ref{app:anemone-background} gives the correspondence). For standalone PCs, the derivatives in \cref{eq:anemone} are nonnegative, so the normalized update preserves nonnegative weights.

Differentiating the normalized model in \cref{eq:Z} reveals why \cref{eq:anemone} does not apply directly.
For directly parameterized sum weights, holding the neural outputs and circuit inputs fixed, let $F_{\x}^+(n,c) := \partial \log\tilde p(\x_0;\params_t,\wparams) / \partial \log \wparams_{n,c}$ and $F_{\x}^-(n,c) := \partial \log Z(\params_t,\wparams) / \partial \log \wparams_{n,c}$. Then
    \begin{align*}
        \frac{\partial \log p_{\vparams,\wparams}(\x_0\mid \x_t)}{\partial \log \wparams_{n,c}} = F_{\x}^+(n,c) - F_{\x}^-(n,c).
    \end{align*}
By construction, both \(F_x^+(n,c)\) and \(F_x^-(n,c)\) are nonnegative because the circuit computes $\tilde{p}$ and $Z$ using only additions and multiplications of nonnegative values; their derivatives with respect to $\log\wparams_{n,c}$ are therefore nonnegative as well. Their difference, however, can be negative. Consequently, directly applying \cref{eq:anemone} no longer guarantees nonnegative circuit weights. To address this problem, we propose a dual-flow multiplicative update that retains the two flows separately and replaces their subtraction with a ratio of positive terms:

\begin{align}
    \wparams^{\mathrm{new}}_{n,c}
    \propto
    \wparams_{n,c} \cdot
    \frac{1+\eta_{n,c} \cdot F_{\mathrm B}^{+}(n,c)}
         {1+\eta_{n,c} \cdot F_{\mathrm B}^{-}(n,c)},
    \label{eq:signed-ratio-update}
\end{align}
where $\eta_{n,c} := \alpha \cdot T_{\vparams,\wparams} (n,c) / \wparams_{n,c}$ and $F_{\mathrm B}^{\pm}$ denote their minibatch averages. The following proposition establishes that the update preserves valid circuit parameters, recovers standard circuit learning in the standalone setting, and agrees locally with the signed-gradient update.

\begin{proposition}[Properties of the dual-flow update]\label{prop:signed-flow-update}
Fix an active sum node with normalized weights $\wparams_{n,c} \!>\! 0$, flows $F_{\mathrm B}^{\pm} (n,c) \!\geq\! 0$,
and scales $0 \!<\! T_{\vparams,\wparams}(n,c) \!<\! \infty$. \cref{eq:signed-ratio-update} satisfies:
\vspace{-1.6em}
\begin{enumerate}
    \renewcommand{\labelenumi}{(\roman{enumi})}
    \setlength{\itemsep}{2pt}
    \setlength{\parskip}{0pt}
    \item For every $\alpha\geq0$, $\wparams^{\mathrm{new}}_{n,c}>0$ and $\sum_{c\in\ch(n)}\wparams^{\mathrm{new}}_{n,c}=1$ even when $F_{\mathrm B}^{+}(n,c) - F_{\mathrm B}^{-}(n,c) < 0$. 
    \item When applied to a standalone PC with normalized inputs and sum weights, the dual-flow update in \cref{eq:signed-ratio-update} reduces to the standard EM update in \cref{eq:anemone}.
    \item Under (ii) and full batch $\mathrm{B}$, \cref{eq:signed-ratio-update} recovers the full-batch EM update as $\alpha \rightarrow \infty$.
    \item As $\alpha \to 0$, \cref{eq:signed-ratio-update} agrees with \cref{eq:anemone} up to $\bigO(\alpha^2)$.
\end{enumerate}
\end{proposition}

The dual-flow update thus recovers standard circuit learning
in the standalone setting and remains valid when conditional
normalization introduces negative gradients. It also matches the normalized signed-gradient update to first order. Together with neural backpropagation, this enables joint training under the conditional denoising objective in \cref{eq:codd-objective}. Appendix~\ref{app:signed-flow-proof} provides the proof.

\section{Instantiating a Context-Parameterized Joint Denoiser}
\label{sec:architecture}

Having established how to jointly train the neural and circuit parameters, we now specify how the neural network uses $\x_t$ to modulate the circuit in \cref{eq:codd2}. A naive parameterization would require the network to output a distinct context-dependent value for every circuit parameter, causing its output dimension to scale with the size of the circuit. Instead, we retain the shared circuit parameters $\wparams$ and use $\params_t=f_{\vparams} (\x_t)$ to modulate both parameterized node types: token potentials modulate the input-node distributions, while a compact set of gates reweight groups of sum-node parameters.

\subsection{Learning Token Potentials}
\label{sec:token-potentials}

We first describe how neural token potentials modulate the circuit's input distributions. Before modulation, each input node $n$ over variable $X_i$ represents a categorical distribution $\pi_n$ over the vocabulary $\mathcal V$, where $\pi_n(v)$ is the probability assigned to token $v$. These probabilities belong to the circuit parameters $\wparams$. Given $\x_t$, the neural network produces $\params_t=f_{\vparams} (\x_t)$, whose token-potential component $\psi_t$ assigns a nonnegative value $\psi_{t,i}(v)$ to each token $v$ for $X_i$. Recall that $g_n$ in \cref{eq:bg-pc} denotes the input distribution over $X_i$. We obtain its modulated form by normalizing the product of the PC input distribution and the neural potentials:
    \begin{equation}
        g_n(v;\params_t,\wparams)=
\frac{\pi_n(v)\,\psi_{t,i}(v)}
{\sum_{v'\in\mathcal V}\pi_n(v')\,\psi_{t,i}(v')}.
\label{eq:token-potentials}
    \end{equation}
\cref{eq:token-potentials} locally normalizes the pointwise product $\pi_n\psi_{t,i}$, making $g_n(\cdot;\params_t,\wparams)$ a categorical distribution over $X_i$. All input nodes over $X_i$ share the same potential vector $\psi_{t,i}$, so the neural network produces one vector per position rather than per input node. Their distinct base distributions $\pi_n$ allow these nodes to retain different token preferences. The local normalizer depends on both the neural potentials and circuit parameters, and is retained in likelihood evaluation and differentiation.

Normalizing \cref{eq:token-potentials} over the entire vocabulary for every input node would be computationally expensive,  even though the top \(1{,}024\)
candidate tokens retain more than \(99\%\) of the backbone's predictive probability mass on average, as quantified in \cref{fig:evidence_truncation} in the appendix.
We therefore restrict computation to candidates favored by the backbone: at each position, we retain the top-$k$ logits, apply a softmax over the retained candidates, and set the remaining potentials to zero. After candidate selection, this reduces each input normalization from $\mathcal O(|\mathcal V|)$ to $\mathcal O(k)$. During training, we ensure that the clean token is retained so that truncation does not force its likelihood to zero. 

Crucially, the local normalization in \cref{eq:token-potentials} fits directly into the dual-flow framework of \cref{sec:joint-learning}. When $\pi_n$ is learned as part of $\wparams$, derivatives through both the numerator and local normalizer contribute to the positive and negative flows, respectively. The same dual-flow update can therefore optimize the input parameters together with the sum parameters. Appendix~\ref{app:training_detail} reports the choice of $k$ and the remaining optimization hyperparameters.

\subsection{Learning Coupling Potentials}
\label{sec:coupling-potentials}

Next, we describe how neural gates can modulate the circuit's
sum weights. In practical PCs, sum parameters are typically arranged in matrices by grouping sum nodes that share the same children, and they determine how the circuit combines its child distributions \citep{liu2024scaling,peharz2020einsum,loconte2025what}.
Sum nodes sharing the same children define a row-normalized weight matrix $\mathbf W$: rows index sum nodes and columns index child distributions. Its entries belong to $\wparams$. We partition $\mathbf W$ into $n\times n$ parameter groups of width $b$, each containing $b\times b$ weights. Given $\x_t$, a compact neural head predicts one gate per group. Exponentiating these outputs yields $\mathbf G_t\in\mathbb R_{>0}^{n\times n}$, with each scalar gate scaling its corresponding parameter group:
\begin{equation}
    \mathbf W_t
    =
    \operatorname{RowNorm}\!\left(
        \mathbf W\odot
        \bigl(\mathbf G_t\otimes\mathbf 1_{b\times b}\bigr)
    \right),
    \label{eq:conditioned-edge}
\end{equation}
where $\odot$ denotes elementwise multiplication, $\otimes$ denotes the Kronecker product, $\mathbf 1_{b\times b}$ is an all-ones matrix, and $\operatorname{RowNorm}$ normalizes each row. Zero log-gates  recover $\mathbf W$, while context-dependent gates adjust the relative mixture weights. Row normalization preserves valid mixtures, and the unchanged circuit structure preserves tractable marginalization and conditioning.

\begin{wrapfigure}{r}{0.5\textwidth}
    \centering
    \vspace{-1.3em}
    \includegraphics[width=\linewidth]{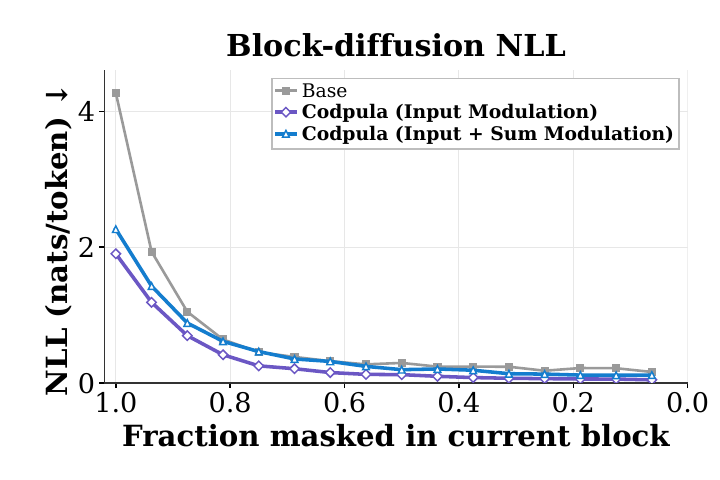}
    \vspace{-2.5em}
    \caption{\textbf{Block-diffusion NLL on Dream.}
    Negative log-likelihood (nats/token) versus the fraction of masked tokens in the current block. Both Codpula variants substantially reduce NLL when most of the block is masked.}
    \label{fig:block_diffusion_nll}
    \vspace{-1.6em}
\end{wrapfigure}

Together, \cref{eq:token-potentials,eq:conditioned-edge} turn the PC into a context-parameterized denoiser. For each $\x_t$, the backbone produces both token potentials that modulate the circuit’s input distributions and group-level gates $\mathbf{G}_t$ that reweight sum parameters. These neural outputs allow the same circuit to express context-dependent token preferences and dependencies through a compact modulation of its shared base parameters. The circuit structure remains fixed, preserving exact marginalization and conditional inference for each $\params_t$. \Cref{fig:block_diffusion_nll} compares the resulting conditional likelihoods under block diffusion. Both variants of Codpula achieve substantially lower NLL than the base model at high masking rates, with the largest improvement when the current block is fully masked.

\section{Decoding from a Learned Joint Denoiser}
\label{sec:decoding}
% \vspace{-0.5em}

Decoding begins from a partially masked sequence. At each denoising step $t$, the learned denoiser defines a tractable joint distribution over the current set of masked variables under consideration, conditioned on the visible tokens. The decoder uses this distribution to decide what to propose and may use its joint structure to decide where to decode, \ie which proposals to commit. Committed values become visible, while the others remain masked for subsequent steps. We first describe how proposals are obtained from the joint distribution, then review standard rules for selecting where to commit, and finally introduce our dependency-aware commitment rule.

To obtain proposals, we use the current neural outputs $\params_t$ to modulate the circuit's input distributions and sum weights as in \cref{eq:token-potentials,eq:conditioned-edge}. Before sampling, we control the sharpness of the neural token preferences by dividing the backbone logits by a temperature $\gamma>0$ before normalization. Writing $\psi_{t,i}$ for the unscaled token potentials, the resulting input distribution is
\begin{align*}
    g_n^{(\gamma)}(v;\params_t,\wparams)
    =
    \frac{
        \pi_n(v)\,\psi_{t,i}(v)^{1/\gamma}
    }{
        \sum_{v'\in\mathcal V}
        \pi_n(v')\,\psi_{t,i}(v')^{1/\gamma}
    }.
    \label{eq:decode-temperature}
\end{align*}
Temperature scaling thus acts on the neural potentials within each input distribution, retaining the learned categorical parameters $\pi_n$ and the current modulated sum weights. We then condition the circuit on the visible tokens and jointly sample the masked variables under consideration.

Once token proposals have been sampled, the decoder decides which positions to commit. The joint denoiser supports standard confidence-, margin-, and entropy-based selection, as well as random selection \citep{nie2025large,ye2025dream,kim2025train}. Our dependency-aware decoder retains these rules for candidate selection, then uses the joint distribution to assess dependencies among the candidates before commitment. Beyond these tokenwise criteria, the learned joint distribution provides conditional queries for adapting which proposals are committed together.

\begin{figure}
    \centering
    \includegraphics[width=\linewidth]{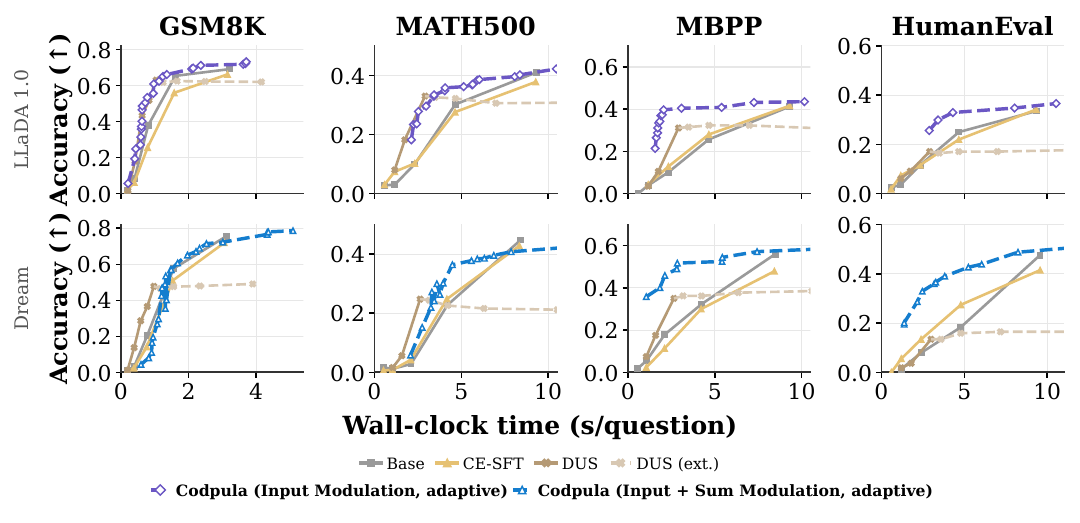}
    \vspace{-24pt}
    \caption{\textbf{Accuracy-latency trade-offs on LLaDA-1.0 and Dream.} Accuracy versus wall-clock time per question for adaptive Codpula, the base backbone, and DUS on math and code benchmarks.}
    \label{fig:wallclock-main}
    % \vspace{-1em}
\end{figure}
We first form proposals for all masked tokens under consideration (\eg those in the current block for block diffusion). For each such token $i$, let $\hat{\x}^{-i}$ denote the proposals at all other positions under consideration. We measure how conditioning on them changes the log-probability of $\hat{x}^{i}$:
\begin{equation}
    \delta_i = \left| \log p_{\vparams,\wparams}\!\left(\hat{x}^{i}\mid \hat{\x}^{-i},\x_t\right) - \log p_{\vparams,\wparams}\!\left(\hat{x}^{i}\mid \x_t\right) \right|.
    \label{eq:dep-score}
\end{equation}

Here, $\delta_i$ is the absolute conditional pointwise mutual information between proposal $i$ and the rest, where smaller values indicate weaker dependence. We commit tokens with $\delta_i<\tau$ ($\tau>0$) and defer the rest, committing the lowest-scoring token if none qualify. Deferred tokens are reconsidered at the next step under the updated context. For $m$ tokens, all scores require one complete-assignment, $m$ singleton, and $m$ leave-one-out marginals. These queries share the visible-evidence normalizer, can be batched with the same neural outputs, and require no additional backbone evaluations. Algorithm~\ref{alg:decode} gives the block-decoding instantiation; Appendix~\ref{app:decoding-details} details the queries and overhead.
% \vspace{-1.4em}

\section{Experiments}
% \vspace{-0.4em}
We evaluate Codpula on mathematical reasoning and code generation using four modern dLLM backbones. We examine its accuracy across parallel-decoding budgets, its accuracy–latency trade-off under adaptive commitment, and the contributions of input and coupling modulation.

\begin{table}[t]
\centering
\caption{\textbf{Parallel Decoding Performance (LLaDA 2.0).} Accuracy (\%) under low-confidence decoding as the number of tokens decoded per diffusion step increases. Parentheses after each benchmark denote the maximum generation length; column headers denote tokens decoded per step. \colorbox{lightgreen}{Shaded} indicates the best and \underline{underline} indicates the second-best performance in each column.}
\label{tab:llada2_tokens_per_step}

\resizebox{0.75\linewidth}{!}{%
\begin{tabular}{ll cccc cccc}
\toprule
\multirow{2}{*}[-0.3em]{\textbf{Model}} & \textbf{Method} & \multicolumn{4}{c}{\textbf{GSM8K (512)}} & \multicolumn{4}{c}{\textbf{MATH500 (1024)}} \\
\cmidrule(lr){3-6} \cmidrule(lr){7-10}
 & {\footnotesize\textit{tokens/step}} & \textbf{2} & \textbf{4} & \textbf{8} & \textbf{16} & \textbf{2} & \textbf{4} & \textbf{8} & \textbf{16} \\
\midrule

\multirow{7}{*}{\shortstack[c]{\textbf{LLaDA}\\\textbf{2.0}}}
 & Base & 70.3 & 68.6 & 36.7 & 2.4 & 65.6 & 55.6 & 28.2 & 1.4 \\
 & ReMDM & 71.0 & 66.6 & 39.0 & 2.4 & 63.2 & 56.8 & 24.4 & 1.4 \\
 & DUS & N/A & N/A & \cellcolor{lightgreen}60.3 & \underline{8.5} & N/A & N/A & \cellcolor{lightgreen}41.8 & 4.4 \\
 & Tolerator & 64.4 & 56.2 & 19.9 & 2.8 & 58.6 & 51.6 & 16.6 & 2.6 \\
 & CE-SFT & \underline{80.7} & \underline{69.2} & 19.9 & 1.1 & \underline{70.0} & 56.6 & 13.8 & 2.0 \\
 & CoDD & 71.6 & 69.0 & 47.3 & 4.2 & 65.4 & \underline{61.6} & 31.8 & \underline{6.8} \\
 & Codpula & \cellcolor{lightgreen} 81.4 & \cellcolor{lightgreen} 75.3 & \underline{57.9} & \cellcolor{lightgreen} 12.7 & \cellcolor{lightgreen} 72.2 & \cellcolor{lightgreen} 62.8 & \underline{41.0} & \cellcolor{lightgreen}12.0 \\
\bottomrule
\end{tabular}%
}
\vspace{-2em}
\end{table}
\textbf{Models and training.}
Our experiments include LLaDA-1.0-8B-Instruct \citep{nie2025large}, LLaDA-2.0-mini \citep{bie2025llada20}, Dream-v0-Instruct-7B \citep{ye2025dream}, and SDAR-8B-Chat-b16 \citep{JetAstra2025}. For LLaDA models, we use two-stage training: we first train the PC with the backbone frozen, then update the backbone through LoRA adapters while keeping the PC fixed. For Dream and SDAR, we jointly optimize the PC and the backbone adapters. The two-stage schedule for LLaDA is motivated by the sensitivity to finetuning among this family.

\textbf{Baselines.}
We use static decoding with the original instruction-tuned backbone as our reference and compare against both training-based and training-free baselines. CE-SFT trains LoRA adapters with masked token cross-entropy, whereas CoDD \citep{li2026breaking} couples the frozen backbone's predictions with a separately trained PC. The training-free baselines include Dynamic Conf. \citep{wu2025fast}, DUS \citep{luxembourg2026plan}, ReMDM \citep{wang2025remasking}, and Tolerator \citep{tian2025finish}, all of which modify only the decoding procedure while keeping the backbone frozen. Appendix~\ref{app:exp_detail} provides supporting evidence for the LLaDA training schedule, full training and decoding configurations, and detailed descriptions of the decoding baselines.

\textbf{Mathematical reasoning.}
We evaluate LLaDA-1.0, LLaDA-2.0, and Dream on MATH500 \citep{math500-lightman2023letsverifystepstep} and GSM8K \citep{gsm8k-cobbe2021training}, using all 500 and 1,319 test problems, respectively. For LLaDA-2.0, we additionally evaluate on AIME 2024 and AIME 2025 \citep{aime}, using combined AIME I and II in each year with chain-of-thought prompting. All mathematical reasoning evaluations are zero-shot and use the lm-evaluation-harness framework \citep{eval-harness}. We report final-answer accuracy, using mathematical equivalence checking for MATH500 and exact match after answer extraction for GSM8K. For AIME, we prioritize the final boxed answer during extraction and report exact-match accuracy after string normalization.

\textbf{Code generation.}
For code generation, LLaDA-1.0 and Dream are evaluated on MBPP \citep{austin2021} and HumanEval \citep{chen2021evaluating}, while SDAR is evaluated on the extended-test variants, MBPP+ and HumanEval+ \citep{liu2023codegen}. The extended test suites provide a more stringent assessment of the functional correctness of generated programs. All code-generation evaluations use zero-shot instruction-style prompts, with performance measured by pass@1. Dataset sizes are summarized in Appendix~\ref{app:tab: eval_size}, with evaluation details provided in the same appendix.

\begin{table}[t]
\centering
\caption{\textbf{Parallel Decoding Performance (SDAR \& Dream).} Accuracy (\%) on HumanEval+ and MBPP+ as the number of tokens decoded per step increases. SDAR uses block size 16, max generation length 512, and low confidence decoding; Dream uses full diffusion with generation length 128 and entropy decoding. \colorbox{lightgreen}{Shaded} indicates the best and \underline{underline} indicates the second-best performance within each model block.}
\label{tab:sdar_results}

\resizebox{0.78\linewidth}{!}{%
\begin{tabular}{ll cccc cccc}
\toprule
\multirow{2}{*}[-0.3em]{\textbf{Model}} & \textbf{Method} & \multicolumn{4}{c}{\textbf{HumanEval+}} & \multicolumn{4}{c}{\textbf{MBPP+}} \\
\cmidrule(lr){3-6} \cmidrule(lr){7-10}
 & {\footnotesize\textit{tokens/step}} & \textbf{1} & \textbf{2} & \textbf{4} & \textbf{8} & \textbf{1} & \textbf{2} & \textbf{4} & \textbf{8} \\
\midrule

% SDAR Block
\multirow{7}{*}{\textbf{SDAR}}
 & Base & 68.9 & 51.2 & 28.1 & 10.4 & 62.2 & 41.5 & 24.6 & 8.5 \\
 & ReMDM & 65.2 & 50.6 & 28.7 & 14.0 & 58.2 & 43.7 & 23.3 & 6.6 \\
 & DUS & N/A & N/A & \underline{37.8} & 12.2 & N/A & N/A & \cellcolor{lightgreen} 35.2 & 12.7 \\
 & CE-SFT & 64.6 & 52.4 & 32.9 & 11.0 & 61.9 & 42.6 & 23.0 & 6.6 \\
 & CoDD & \underline{72.6} &  \underline{55.5} & 32.9 & 11.6 & 65.1 & 42.9 & 25.1 & 9.8 \\
 & Codpula & \underline{72.6} & \cellcolor{lightgreen}59.8 & 36.0 & \underline{17.1} & \underline{67.7} & \cellcolor{lightgreen} 55.6 & \underline{34.4} & \cellcolor{lightgreen} 18.5 \\
 & Codpula (Both) & \cellcolor{lightgreen} 73.8 & \cellcolor{lightgreen}59.8 & \cellcolor{lightgreen} 38.4 & \cellcolor{lightgreen} 17.7 & \cellcolor{lightgreen} 69.3 & \underline{52.9} & \underline{34.4} & \underline{18.3} \\
\midrule

% Dream Block
\multirow{4}{*}{\textbf{Dream}}
 & Base & 39.6 & 34.1 & 26.8 & 24.4 & 48.1 & 46.0 & 38.9 & 31.2 \\
 & CE-SFT & \underline{44.5} & \underline{39.0} & 26.8 & 25.0 & 48.4 & 43.4 & 30.4 & 13.0 \\
 & Codpula & 33.5 & 34.8 & \underline{32.9} & \cellcolor{lightgreen} 31.7 & \underline{53.4} & \cellcolor{lightgreen} 51.6 & \underline{45.2} & \cellcolor{lightgreen} 35.7 \\
 & Codpula (Both) & \cellcolor{lightgreen} 45.7 & \cellcolor{lightgreen} 45.1 & \cellcolor{lightgreen} 41.5 & \underline{29.9} & \cellcolor{lightgreen} 59.0 & \underline{49.2} & \cellcolor{lightgreen} 47.9 & \underline{33.3} \\
\bottomrule
\end{tabular}%
}
% \vspace{-1em}
\end{table}

\begin{figure}
    \centering
    \includegraphics[width=\linewidth]{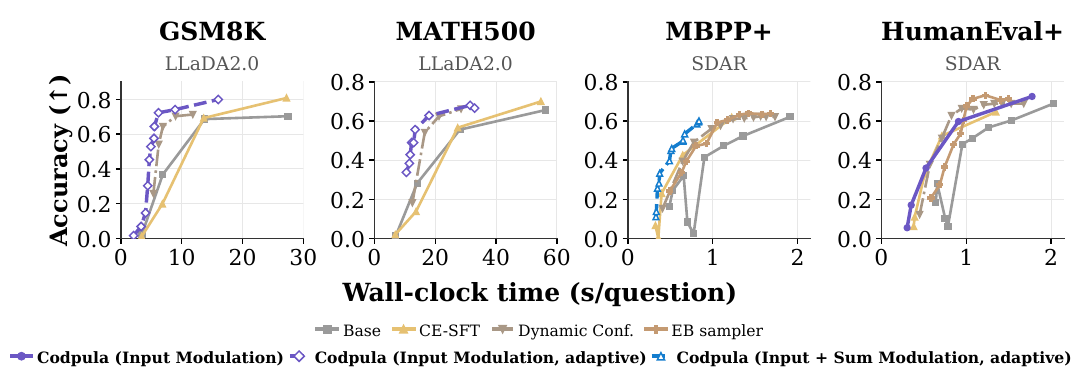}
    \vspace{-24pt}
    \caption{\textbf{Accuracy-latency trade-offs on LLaDA-2.0 and SDAR.} Accuracy versus wall-clock time per question for Codpula with uniform and adaptive decoding, base backbone, dynamic-confidence, and EB-sampler decoding on math (LLaDA-2.0) and code (SDAR) benchmarks.}
    \label{fig:wallclock-llada2-sdar}
    % \vspace{-1.5em}
\end{figure}

\textbf{Training through the joint model improves parallel prediction.}
On LLaDA-2.0 GSM8K (\cref{tab:llada2_tokens_per_step}), CE-SFT nearly matches
Codpula at two tokens per step but falls below the backbone
at eight. Codpula improves over the backbone in both
settings, by $11.1$ and $21.2$ percentage points,
respectively. Thus, improvements from tokenwise
fine-tuning do not necessarily translate to effective
parallel prediction. The ablation in
\cref{tab:coding_ablation} supports training through the circuit: backbone adaptation adds gains over fitted circuit in seven of ten settings.

\textbf{Context-dependent coupling provides additional,
task-dependent gains.}
Input-only Codpula already improves over the backbone
across multiple settings
(\cref{tab:llada2_tokens_per_step,tab:merged_results}).
Sum modulation provides further gains, most clearly
on Dream's full-sequence HumanEval+ evaluation
(\cref{tab:sdar_results}): it improves the input-only
variant at one to four tokens per step, including a
setting where that variant falls below the backbone. These results indicate that
adapting mixture weights can improve on token
reweighting alone, with benefits that depend on the
task and decoding regime.

\textbf{The gains extend to measured generation latency.}
Figures~\ref{fig:wallclock-main}
and~\ref{fig:wallclock-llada2-sdar} show improved
accuracy--latency trade-offs after accounting for circuit
computation, with adaptive decoding improving selected
operating points. This improvement comes despite
additional computation: on LLaDA-1.0/MATH500, adaptive
Codpula incurs $6.8$--$8.2\%$ overhead at matched
numbers of backbone evaluations
(\cref{tab:codpula-overhead}).
Under EOS-based stopping, latency also reflects
generation length; the shorter generations in
\cref{tab:aime-lowconf-generation} therefore contribute
to the observed speedups
(Appendix~\ref{app:early_stop}).

\begin{table*}[t]
\centering
\caption{\textbf{Parallel Decoding Comparison across Math and Code Benchmarks (LLaDA-1.0 \& Dream).} Accuracy (\%) as the number of tokens decoded per step increases. Parentheses after each benchmark denote the maximum generation length; column headers denote tokens decoded per step. Math benchmarks use low-confidence decoding; for code, LLaDA uses margin decoding on MBPP and low-confidence decoding on HumanEval, while Dream uses entropy decoding. \colorbox{lightgreen}{Shaded} indicates the best and \underline{underline} indicates the second-best performance within each model block.}
\label{tab:merged_results}

\resizebox{0.95\textwidth}{!}{%
\begin{tabular}{ll ccc ccc ccc ccc}
\toprule
\multirow{2}{*}[-0.3em]{\textbf{Model}} & \textbf{Method} & \multicolumn{3}{c}{\textbf{GSM8K (256)}} & \multicolumn{3}{c}{\textbf{MATH500 (512)}} & \multicolumn{3}{c}{\textbf{MBPP (512)}} & \multicolumn{3}{c}{\textbf{HumanEval (512)}} \\
\cmidrule(lr){3-5} \cmidrule(lr){6-8} \cmidrule(lr){9-11} \cmidrule(lr){12-14}
 & {\footnotesize\textit{tokens/step}} & \textbf{2} & \textbf{4} & \textbf{8} & \textbf{2} & \textbf{4} & \textbf{8} & \textbf{2} & \textbf{4} & \textbf{8} & \textbf{2} & \textbf{4} & \textbf{8} \\
\midrule

% LLaDA Block
\multirow{5}{*}{\shortstack[c]{\textbf{LLaDA}\\\textbf{1.0}}}
 & Base & 69.1 & 65.4 & 37.8 & \underline{41.0} & 30.2 & 10.0 & 41.3 & 25.7 & 10.1 & 33.5 & \underline{25.0} & 11.6 \\
 & CE-SFT & 66.2 & 56.0 & 25.6 & 37.8 & 27.6 & 10.2 & 41.6 & 28.0 & \underline{12.8} & 34.2 & 22.0 & 11.6 \\
 & ReMDM & 69.6 & 64.3 & 37.2 & 40.2 & 31.8 & 11.2 & 38.1 & 24.9 & 8.2 & \underline{34.8} & \underline{25.0} & 12.2 \\
 & CoDD & \underline{71.5} & \underline{66.5} & \underline{50.3} & 38.6 & \underline{34.8} & \cellcolor{lightgreen} 22.8 & \underline{43.2} & \underline{29.2} & 9.7 & \underline{34.8} & \underline{25.0} & \underline{12.8} \\
 & Codpula & \cellcolor{lightgreen} 71.7 & \cellcolor{lightgreen} 69.0 & \cellcolor{lightgreen} 52.4 & \cellcolor{lightgreen} 41.6 & \cellcolor{lightgreen} 35.2 & \underline{20.8} & \cellcolor{lightgreen} 48.3 & \cellcolor{lightgreen} 31.9 & \cellcolor{lightgreen} 15.6 & \cellcolor{lightgreen} 37.2 & \cellcolor{lightgreen} 29.3 & \cellcolor{lightgreen} 16.5 \\
\midrule

% Dream Block
\multirow{6}{*}{\textbf{Dream}}
 & Base & 75.5 & 57.6 & 20.5 & 44.8 & 22.8 & 2.8 & 56.0 & 32.3 & 17.9 & \underline{47.6} & 18.3 & 7.9 \\
 & CE-SFT & 71.9 & 50.9 & 14.1 & 42.8 & 24.8 & 4.2 & 47.9 & 30.0 & 11.3 & 41.5 & 27.4 & 13.4 \\
 & ReMDM & 74.0 & 56.7 & 21.3 & 42.2 & 20.0 & 3.0 & 56.0 & 40.1 & 26.1 & 47.0 & 28.7 & 12.2 \\
 & CoDD & \underline{77.9} & 58.6 & 14.8 & \underline{46.4} & 29.0 & 6.2 & 58.0 & \underline{43.2} & \underline{34.6} & 42.1 & 31.7 & 18.3 \\
 & Codpula & 77.2 & \underline{60.9} & \underline{23.0} & \underline{46.4} & \cellcolor{lightgreen} 31.8 & \underline{8.4} & \cellcolor{lightgreen} 60.3 & \cellcolor{lightgreen} 48.6 & 31.5 & \cellcolor{lightgreen} 48.8 & \cellcolor{lightgreen} 38.4 & \cellcolor{lightgreen} 22.0 \\
 & Codpula (Both) & \cellcolor{lightgreen} 78.8 & \cellcolor{lightgreen} 64.9 & \cellcolor{lightgreen} 24.9 & \cellcolor{lightgreen} 48.0 & \underline{30.4} & \cellcolor{lightgreen} 11.4 & \underline{59.1} & \cellcolor{lightgreen} 48.6 & \cellcolor{lightgreen} 36.6 & 45.7 & \underline{37.2} & \underline{20.7} \\
\bottomrule
\end{tabular}%
}
\end{table*}
% \vspace{-0.5em}
\section{Related Work and Conclusion}
\label{sec:related-work-conclusion}
% \vspace{-0.5em}
\paragraph{Joint prediction and adaptive decoding.}
Masked diffusion models, including MDLM and MD4
\citep{sahoo2024simple,shi2024simplified}, LLaDA and Dream
\citep{nie2025large,ye2025dream}, and block-diffusion models
\citep{arriola2025block}, typically factorize predictions
across positions, omitting dependencies among parallel
outputs. Remedies include auxiliary copulas
\citep{liu2025discrete}, sequence-level energy corrections
\citep{xu2025energy}, and mixture-based distillation
\citep{hayakawa2025distillation}.
CoDD couples a frozen denoiser with a learned circuit prior
\citep{li2026breaking}, while tensor-train modeling predicts
context-conditioned joint distributions through a modified
output head \citep{kim2026tensortrainjointmodelingfewstep}.
Complementary decoding strategies improve token ordering
\citep{kim2025train}, adapt parallelism using confidence
or entropy \citep{wu2025fast,ben-hamu2025accelerated},
assess parallel proposals with an auxiliary autoregressive
model \citep{israel2025accelerating}, or remask generated
tokens \citep{wang2025remasking}.
We jointly learn token potentials and context-dependent
coupling under a conditional objective and query the
resulting joint distribution for adaptive commitment.

\paragraph{Probabilistic circuits and their applications.}
Probabilistic circuits represent expressive joint distributions
with tractable marginalization and conditioning
\citep{poon2011sum,ProbCirc20}.
Neural extensions use network outputs to parameterize
context-dependent distributions
\citep{shao2020conditional,
yidouweng2026learningtractabledistributionslanguage,
grivas2026fastexpressivemultibyteprediction}.
PCs and related models also guide constrained autoregressive
generation \citep{zhang2023tractable,zhang2024adaptable}
and diffusion-based image inpainting \citep{liu2024image}.
We integrate the circuit into the denoiser, modulate its
coupling parameters with backbone outputs, and train the
neural components through it.
Building on Anemone's minibatch EM update
\citep{liu2025rethinkingprobabilisticcircuitparameter},
we develop a multiplicative update that accommodates
signed flows from conditional normalization while
preserving valid circuit parameters.

In this work, we introduce Codpula, a context-parameterized joint denoiser that learns token predictions and their coupling while retaining exact normalization and conditional inference. To support conditional training, we derive a dual-flow update, and the learned joint distribution enables dependence-adaptive decoding. Experiments on mathematical reasoning and code generation show improved accuracy-latency trade-offs across multiple backbones. Additional circuit computation and fixed-size coupling windows remain as limitations, motivating more efficient inference and joint modeling over longer sequences.

\section*{Acknowledgments}
This work was funded in part by the National University of Singapore under its Start-up Grant (Award No: SUG-251RES2505); this work was also funded in part by the DARPA CLARA program, and gifts from Cisco Research, Qualcomm, and Amazon.
HZ and ZZ were supported in part by the DARPA CLARA program under Agreement No. HR0011269E227.

\bibliography{iclr2027_conference}
\bibliographystyle{iclr2027_conference}

\clearpage
\appendix
\section{Experiment Details}
\label{app:exp_detail}
\subsection{Training}
\label{app:training_detail}

\begin{table*}[h]
    \centering
    \small
    \setlength{\tabcolsep}{5pt}
    \caption{Backbone models used in our experiments. Parameter counts are the nominal sizes reported by the model developers; for dense models, all parameters are active during each forward pass. Block denotes the decoding block length used in our experiments.}
    \label{tab:model_summary}
    \resizebox{\textwidth}{!}{
    \begin{tabular}{@{}lrrrlll@{}}
        \toprule
        Model & Total params. & Active params. & Layers & Architecture & Diffusion structure & Block \\
        \midrule
        LLaDA-1.0-8B-Instruct \citep{nie2025large}
            & 8B  & 8B   & 32 & Dense & Full-sequence masked diffusion & 32 \\
        LLaDA-2.0-mini \citep{bie2025llada20}
            & 16B & 1.4B & 20 & MoE   & Block-causal masked diffusion & 32 \\
        Dream-v0-Instruct-7B \citep{ye2025dream}
            & 7B  & 7B   & 32 & Dense & Full-sequence masked diffusion & 32 \\
        SDAR-8B-Chat-b16 \citep{JetAstra2025}
            & 8B  & 8B   & 36 & Dense & Semi-autoregressive block diffusion & 16 \\
        \bottomrule
    \end{tabular}}
\end{table*}

\paragraph{SFT on LLaDA.}
Continued fine-tuning does not consistently improve instruction-tuned LLaDA models, especially on mathematical ability. In d1, supervised fine-tuning improves over LLaDA-8B-Instruct in only seven of twelve evaluated settings \citep{zhao2025d}. On LLaDA-1.5, DSFT reports that standard SFT offers negligible or even negative effects on the downstream tasks \citep{chen2025dsftinspiringdiffusionlarge}. Similarly, \cite{sun2026blockwise} report a decrease on MATH from $31.7\%$ to $29.6\%$ under classical SFT. CDLM provides a similar observation: its initial LLaDA-1.0 training on approximately 7500 Bespoke prompts reduces GSM8K accuracy from $77.1\%$
to $72.1\%$ \citep{kim2026cdlmconsistencydiffusionlanguage}. Taken together, these findings motivate treating the backbone update schedule as a consequential design choice. We therefore separate circuit training from backbone fitting for LLaDA model family and retain an independently tuned CE-SFT baseline in the comparison. 

\paragraph{Training data and masking.} For each sequence, we sample a masking probability $t$ uniformly from $[0,1)$ and independently mask response tokens with probability $t$. Prompt tokens remain visible, and padding positions are excluded from the supervised region. The backbone processes the resulting masked sequence; its predictions over the response are then divided into 32-token blocks for circuit training. When the response length is not divisible by 32, we add a final, right-aligned block covering the last 32 response tokens. This block overlaps the preceding block and preserves the response ending without introducing additional padding targets. Responses shorter than a complete block are excluded from circuit training.
\begin{itemize}
\item Mathematical Reasoning.
For LLaDA-1.0, both training stages use the same subset of 1,910 hard problems from NuminaMath \citep{numina_math_datasets}, selected according to the frozen backbone's denoising loss. LLaDA-2.0 uses the full NuminaMath dataset in both stages, while Dream uses the same full dataset during joint training. 

\item Code Generation.
All models use the same training mixture constructed from JinaAI Code Exercises \citep{jinaai2023codeexercises}, KodCode-V1 \citep{xu2025kodcode}, CodeFeedback-Filtered-Instruction \citep{zheng-etal-2024-opencodeinterpreter}, and the non-test splits of MBPP \citep{austin2021}. The mixture contains 98,660 examples in HumanEval- and MBPP-style formats and is used in both training stages for LLaDA-1.0 and during joint training for Dream and SDAR.
\end{itemize}

\paragraph{PC configurations.}
We use homogeneous hidden Markov models with $1{,}024$ latent states per position, compiled into probabilistic circuits. The circuit sequence length is 32 for LLaDA-1.0, LLaDA-2.0, and Dream, and 16 for SDAR. Transition parameters are shared across positions, and categorical leaf distributions are parameterized over the full vocabulary of the corresponding backbone. We train separate circuits for each backbone and task family. Within each backbone and task family, CoDD and Codpula use the same circuit architecture. Parameter conditioning modulates the circuit's sum weights while preserving its topology. Training hyperparameters are summarized in \cref{tab:training-setup}. Circuit training retains the backbone's top-$1{,}024$ evidence candidates per position; an ablation on LLaDA-1.0 examines this choice and its compatibility with full-vocabulary decoding in \cref{app:evidence_truncation}.

\clearpage
\begin{table*}[h]
\centering
\footnotesize
\setlength{\tabcolsep}{2pt}
\caption{Codpula training configurations by backbone and task, including
parameter conditioning. Two-stage runs first train the PC with the
backbone frozen and then train the adapter with the PC fixed. Joint runs
optimize the PC and adapter together, and additionally optimize the
router for Codpula (Both). Router settings for Input Only and Both are
listed separately; all other SDAR settings are shared between the two
variants. A dash in a router row indicates that no router is trained.
Paired stage-specific entries denote stage-1/stage-2 values, and
accumulation thresholds count response blocks. An EM schedule $f$--$e$
remains flat for $f$ updates and then decays cosinely through update $e$.
}
\label{tab:training-setup}
\begin{tabular*}{\linewidth}{@{\extracolsep{\fill}}lcccccc@{}}
\toprule
 & \multicolumn{2}{c}{LLaDA-1.0-8B} & LLaDA2.0-mini & SDAR-8B
 & \multicolumn{2}{c}{Dream-v0-7B} \\
\cmidrule(lr){2-3}\cmidrule(lr){4-4}\cmidrule(lr){5-5}\cmidrule(l){6-7}
 & Math & Code & Math & Code (b16) & Math & Code \\
\midrule
\multicolumn{7}{@{}l}{\emph{Data and objective}}\\
Training scheme          & Two-stage & Two-stage & Two-stage & Joint & Joint & Joint \\
Training examples        & 1,910 & 98,660 & 521,776 & 101,648 & 521,776 & 98,660 \\
Block accum.\ threshold  & 64 & 64 & 400 & 32 & 400 & 400 \\
Inv.-mask weighting      & $10^{-2}$ & $10^{-2}$ & $10^{-2}$ & No & $10^{-2}$ & $10^{-2}$ \\
\addlinespace
\multicolumn{7}{@{}l}{\emph{Circuit}}\\
Block length / latents   & 32\,/\,1,024 & 32\,/\,1,024 & 32\,/\,1,024 & 16\,/\,1,024 & 32\,/\,1,024 & 32\,/\,1,024 \\
Evidence candidates      & 1,024 & 1,024 & 1,024 & All & 1,024 & 1,024 \\
EM step size             & $0.3{\to}0.03$ & $0.3{\to}0.03$ & $0.3{\to}0.03$ & 0.05 & $0.2{\to}0.1$ & $0.2{\to}0.1$ \\
EM schedule              & 67--673 & 120--1,200 & 1,086--end & 1,000--end$^\dagger$ & 1,000--10,000 & 105--1,050\\
\addlinespace
\multicolumn{7}{@{}l}{\emph{Parameter conditioning}}\\
Gate grid (Both)         & --- & --- & --- & $32{\times}32$ & $32{\times}32$ & $32{\times}32$ \\
Router LR (Input Only)   & --- & --- & --- & 0 & --- & --- \\
Router LR (Both)         & --- & --- & --- & $10^{-3}$ & $10^{-3}$ & $10^{-3}$ \\
\addlinespace
\multicolumn{7}{@{}l}{\emph{Adapter}}\\
Learning rate            & $2{\times}10^{-4}$ & $10^{-5}$ & $10^{-6}{\to}10^{-7}$\textsuperscript{$\ddagger$} & $10^{-5}$ & $10^{-6}$ & $10^{-6}{\to}0$ \\
Weight decay             & 0.1 & 0.1 & 0.1 & 0.01 & 0.01 & 0 \\
LoRA $r$/$\alpha$/drop.  & 512/128/0.05 & 128/256/0.05 & 128/256/0 & 128/64/0 & 128/256/0 & 128/256/0 \\
LoRA scaling $\alpha/r$ & 0.25 & 2 & 2 & 0.5 & 2 & 2 \\
Adapted modules          & A & A & B & C & D & D \\
\addlinespace
\multicolumn{7}{@{}l}{\emph{Budget}}\\
Training updates         & 896\,/\,100 & 2,000\,/\,100 & 10,860\,/\,1,540 & 3,000 & 10,847 & 1,050 \\
Ckpt / val.\ interval    & 224\,/\,25 & 400\,/\,25 & 1,350\,/\,385 & 250 & 200 & 150 \\
Selected checkpoint      & 896\,/\,75 & 2,000\,/\,75 & 2,700\,/\,770 & 250 & 5,000 & 1,050 \\
\bottomrule
\end{tabular*}
\begin{minipage}{\textwidth}
\scriptsize
$^\dagger$The selected checkpoint precedes the decay;
$^\ddagger$the adapter learning rate remains flat for 154 updates before
decaying. Module groups A--D are defined below.
\raggedright

\end{minipage}
\end{table*}

\paragraph{Shared training settings.}
All Codpula runs use Anemone EM with momentum 0.9 and evidence temperature 1.0. Adapters are optimized with AdamW without bias adaptation, rsLoRA, or an auxiliary cross-entropy loss. Each GPU processes two sequences per microbatch; padding positions are unsupervised, and gradients on observed evidence are masked. Among the configurations in \cref{tab:training-setup}, only LLaDA2.0 uses gradient clipping, with a maximum norm of 1.0. Parameter conditioning is enabled in Codpula (Both) by training the router at a learning rate of $10^{-3}$ for SDAR and for both Dream task families. For SDAR, Codpula (Input Only) keeps the router fixed at its uniform initialization with learning rate zero; all other hyperparameters are identical to those of Codpula (Both). Module group A adapts per-block query, key, value, and feed-forward output projections, together with the vocabulary head and token embeddings; B adapts fused QKV and the shared-expert MLP; C adapts all attention and MLP projections; and D adapts only query, key, and value projections.

\paragraph{Baseline training and hyperparameter selection.}
CE-SFT trains an independent LoRA adapter with the masked-token cross-entropy objective, while CoDD trains a circuit with the backbone frozen. We tune the baselines independently for each backbone and task family. For CE-SFT, the search covers the adapter learning rate, effective batch size, LoRA rank $r$, scaling ratio $\alpha/r$, and number of training updates. For CoDD, we vary the EM step size, response-block accumulation threshold, and training duration. Intermediate checkpoints are evaluated by downstream generation accuracy, and we report the best-performing configuration for each baseline. The selected baseline configurations are summarized in \cref{tab:baseline-setup}; their learning rates, batch sizes, and LoRA settings are selected independently of Codpula's. Sequence batch size and response-block accumulation are distinct quantities and are reported in their respective units.

\begin{table*}[h]
\centering
\footnotesize
\setlength{\tabcolsep}{4pt}
\caption{Baseline training configurations by backbone and task. CE-SFT trains only the LoRA adapter using masked-diffusion cross-entropy, whereas CoDD trains only the PC with the backbone frozen. LoRA configurations report rank/scale/dropout, and PC configurations report block length $\times$ latent states. Batch sizes count sequences for CE-SFT and response blocks for CoDD.}
\label{tab:baseline-setup}
% \resizebox{\textwidth}{!}{%
\begin{tabular*}{\textwidth}{@{\extracolsep{\fill}}llllrrr@{}}
\toprule
Arm & Trained component & LR / EM step size & Schedule & Batch & Updates & Selected \\
\midrule
\multicolumn{7}{@{}l}{\emph{LLaDA-1.0-8B-Instruct, math} \quad numina\_hard, 1,910 examples}\\
CE-SFT                    & LoRA 512/128/0.05      & $2{\times}10^{-4}$    & Constant   & 64    & 673   & 112 \\
CoDD$^{\dagger\S}$    & PC $32{\times}1{,}024$ & $0.1{\to}0.01$        & 67--end    & 64    & 673   & final \\
\addlinespace
\multicolumn{7}{@{}l}{\emph{LLaDA-1.0-8B-Instruct, code} \quad code\_gen, 98,660 examples}\\
CE-SFT                    & LoRA 128/128/0.05      & $10^{-5}$             & Constant   & 64    & 100   & 75 \\
CoDD$^\S$             & PC $32{\times}1{,}024$ & $0.3{\to}0.03$        & 120--end   & 64    & 673   & final \\
\addlinespace
\multicolumn{7}{@{}l}{\emph{LLaDA2.0-mini, math} \quad numina\_cot, 521,776 examples}\\
CE-SFT                    & LoRA 128/256/0         & $10^{-6}{\to}10^{-7}$ & 154--end   & 48    & 1,540 & 385 \\
CoDD$^\S$       & {PC $32{\times}1{,}024$} &  $0.3{\to}0.03$ & 1,086–10,860         &400      &  6,750    & 1,350 \\
\addlinespace
\multicolumn{7}{@{}l}{\emph{SDAR-8B-Chat b16, code} \quad code\_gen\_sdar, 101,648 examples}\\
CE-SFT                    & LoRA 128/256/0.05      & $10^{-5}$             & Warm-up 20 & 32    & 800   & 700 \\
CoDD$^{\S\ddagger\P}$ & PC $16{\times}1{,}024$ & $0.05{\to}0.005$      & 80--end    & 2 seq.\ ($\approx$23 blk.) & 1,600 & final \\
\addlinespace
\multicolumn{7}{@{}l}{\emph{Dream-v0-7B-Instruct, math} \quad numina\_cot, 521,776 examples}\\
  CE-SFT             & LoRA 128/256/0        & $10^{-5}$    & Constant        & 48     & 10,800 & 5,600 \\
  CoDD           & PC $32{\times}1{,}024$ & $0.2{\to}0.1$ & 5,100--51,000  & 16,384 & 49,527 &  40,612 \\
\addlinespace
\multicolumn{7}{@{}l}{\emph{Dream-v0-7B-Instruct, code} \quad code\_gen, 98,660 examples}\\
CE-SFT                    & LoRA 128/256/0            & $10^{-5}$     & Warm-up 40 &  96 & 1{,}060 & 300 \\
CoDD$^\S$             & PC $32{\times}1{,}024$    & $0.2{\to}0.1$ & 585--5{,}850           & 16{,}384    & 5{,}850          & 5{,}703 \\
\bottomrule
\end{tabular*}
\begin{minipage}{\textwidth}
\scriptsize
\raggedright
$^\dagger$The corpus uses three precomputed mask draws per example.
$^\S$External emissions are constant across latent states, so the evidence cancels from the latent posterior and the M-step fits the prior by exact maximum likelihood.
$^\ddagger$This run uses linear decay and updates EM once per two-sequence precomputed batch, corresponding to approximately 23 response blocks per update.
$^\P$The PC is trained on 3,200 sequences, corresponding to half of the 6,400-sequence precomputed cache.
CE-SFT hyperparameters are selected independently and matched to the corresponding Codpula configuration by corpus, training budget, and learning rate.
\end{minipage}
\end{table*}

\clearpage
\subsection{Evaluation}
\begin{table}[h]
\label{app:tab: eval_size}
    \centering
    \small
    \caption{Datasets and evaluation-set sizes.}
    \label{tab:dataset_statistics}
    \begin{tabular}{llr}
        \toprule
        Dataset & Evaluated split & \# Samples \\
        \midrule
        \multicolumn{3}{l}{\textit{Mathematical reasoning}} \\
        MATH500     & Test                  & 500   \\
        GSM8K       & Test                  & 1,319 \\
        AIME 2024   & AIME I and AIME II    & 30    \\
        AIME 2025   & AIME I and AIME II    & 30    \\
        \midrule
        \multicolumn{3}{l}{\textit{Code generation}} \\
        HumanEval   & Test                  & 164   \\
        MBPP        & Sanitized test        & 257   \\
        HumanEval+  & EvalPlus test         & 164   \\
        MBPP+       & EvalPlus test         & 378   \\
        \bottomrule
    \end{tabular}
\end{table}

\begin{table*}[h]
\centering
\small
\setlength{\tabcolsep}{4pt}
\caption{Evaluation protocol. Sets are always evaluated in full.}
\label{tab:setup-eval}
\resizebox{\textwidth}{!}{
\begin{tabular}{@{}llrllll@{}}
\toprule
Benchmark & Model & $n$ & Metric key & Shots & System prompt & Canvas \\
\midrule
GSM8K            & LLaDA-8B, Dream-7B & 1{,}319 & \texttt{exact\_match,flexible-extract} & 0 & none         & 256 \\
                 & LLaDA2.0-mini      & 1{,}319 & \texttt{exact\_match,flexible-extract} & 0 & none         & 512 \\
MATH500          & LLaDA-8B, Dream-7B & 500     & \texttt{math\_verify,none}             & 0 & task default & 512 \\
                 & LLaDA2.0-mini      & 500     & \texttt{math\_verify,none}             & 0 & task default & 1{,}024 \\
MBPP (sanitized) & LLaDA-8B, Dream-7B & 257     & \texttt{pass\_at\_1,extract\_code}     & 0 & task default & 512 \\
HumanEval        & LLaDA-8B, Dream-7B & 164     & \texttt{pass@1,create\_test}           & 0 & task default & 512 \\
MBPP+            & SDAR-8B-Chat-b16   & 378     & \texttt{pass\_at\_1,extract\_code}     & 0 & task default & 512 \\
HumanEval+       & SDAR-8B-Chat-b16   & 164     & \texttt{pass@1,create\_test}           & 0 & task default & 512 \\
AIME24 / AIME25  & LLaDA2.0-mini      & 30 each & \texttt{exact\_match,none}             & 0 & task default & 10{,}240 \\
\bottomrule
\end{tabular}
}
\end{table*}

\subsection{Decoding}
\subsubsection{Decoding rules chosen for the base}

\begin{table*}[t]
\centering
\caption{\textbf{Alternative Decoding Rules (LLaDA-1.0).}
Accuracy (\%) under alternative token-selection rules, complementing the main-text results. Parentheses after benchmark names denote maximum generation lengths; column headers denote tokens decoded per step.
$^{\dagger}$For MBPP, the second block uses low-confidence decoding; the other benchmarks use margin decoding. Cells marked -- were not run.
\colorbox{lightgreen}{Shading} and \underline{underlining} indicate the best and second-best accuracy, respectively, in each column within each decoding block. \\}
\label{tab:llada_decoding_rules}

\resizebox{\textwidth}{!}{%
\begin{tabular}{ll ccccc cccc cccc ccccc}
\toprule
\multirow{2}{*}[-0.3em]{\textbf{Decoding}} & \textbf{Method} & \multicolumn{5}{c}{\textbf{GSM8K (256)}} & \multicolumn{4}{c}{\textbf{MATH500 (512)}} & \multicolumn{4}{c}{\textbf{MBPP (512)}$^{\dagger}$} & \multicolumn{5}{c}{\textbf{HumanEval (512)}} \\
\cmidrule(lr){3-7} \cmidrule(lr){8-11} \cmidrule(lr){12-15} \cmidrule(lr){16-20}
 & {\footnotesize\textit{tokens/step}} & \textbf{2} & \textbf{4} & \textbf{8} & \textbf{16} & \textbf{32} & \textbf{2} & \textbf{4} & \textbf{8} & \textbf{16} & \textbf{2} & \textbf{4} & \textbf{8} & \textbf{16} & \textbf{2} & \textbf{4} & \textbf{8} & \textbf{16} & \textbf{32} \\
\midrule

% HumanEval source columns: 256, 128, 64, 32, 16 total denoising steps.
% At generation length 512, these map to 2, 4, 8, 16, 32 tokens per step.
% Codpula temperatures: Random 0.05, 0.3, 0.3, --, --; Margin 0.3, 0.3, 0.3, 0.1, 0.05.
\multirow{4}{*}{\textbf{Random}}
 & Base & 57.5 & 42.9 & 22.1 & \underline{4.8} & \underline{1.7} & \cellcolor{lightgreen} 30.8 & 16.2 & 7.6 & 4.0 & 24.1 & 12.5 & \underline{3.9} & \underline{1.6} & \underline{18.9} & 8.5 & \underline{6.1} & -- & -- \\
 & CE-SFT & 52.2 & 41.1 & 21.2 & -- & -- & 23.4 & 17.8 & 9.6 & 7.6 & 24.9 & \underline{13.6} & -- & -- & 17.7 & 9.8 & -- & -- & -- \\
 & CoDD & \underline{58.5} & \underline{45.5} & \underline{24.4} & -- & -- & \underline{26.6} & \cellcolor{lightgreen} 21.4 & \cellcolor{lightgreen} 14.0 & \underline{8.4} & \cellcolor{lightgreen} 27.6 & 12.5 & -- & -- & 17.7 & \underline{12.2} & -- & -- & -- \\
 & Codpula & \cellcolor{lightgreen} 60.1 & \cellcolor{lightgreen} 49.0 & \cellcolor{lightgreen} 27.9 & \cellcolor{lightgreen} 12.3 & \cellcolor{lightgreen} 5.6 & 25.4 & \underline{19.6} & \underline{12.2} & \cellcolor{lightgreen} 9.0 & \underline{26.1} & \cellcolor{lightgreen} 17.1 & \cellcolor{lightgreen} 9.7 & \cellcolor{lightgreen} 4.7 & \cellcolor{lightgreen} 21.3 & \cellcolor{lightgreen} 12.8 & \cellcolor{lightgreen} 8.5 & -- & -- \\
\midrule

\multirow{4}{*}{\textbf{Margin}}
 & Base & \cellcolor{lightgreen} 70.4 & \cellcolor{lightgreen} 64.4 & \cellcolor{lightgreen} 44.3 & \underline{10.1} & \underline{1.7} & \cellcolor{lightgreen} 39.2 & \cellcolor{lightgreen} 31.8 & 18.0 & 4.6 & 40.9 & 24.5 & 9.0 & \underline{2.0} & \underline{35.4} & \underline{25.0} & 11.0 & 4.9 & 2.4 \\
 & CE-SFT & -- & -- & -- & -- & -- & \underline{38.8} & 27.8 & 13.0 & 7.8 & 40.5 & 28.4 & 9.0 & 1.2 & \cellcolor{lightgreen} 36.6 & 24.4 & 14.0 & \cellcolor{lightgreen} 6.1 & 1.8 \\
 & CoDD & -- & -- & -- & -- & -- & 38.6 & \cellcolor{lightgreen} 31.8 & \underline{18.6} & \cellcolor{lightgreen} 9.8 & \cellcolor{lightgreen} 44.8 & \underline{28.8} & \cellcolor{lightgreen} 18.3 & \cellcolor{lightgreen} 7.4 & 29.9 & 21.3 & \underline{14.6} & \cellcolor{lightgreen} 6.1 & \underline{3.1} \\
 & Codpula & \underline{70.1} & \underline{63.1} & \underline{44.2} & \cellcolor{lightgreen} 15.7 & \cellcolor{lightgreen} 5.6 & \underline{38.8} & \underline{31.2} & \cellcolor{lightgreen} 19.4 & \underline{9.6} & \underline{42.4} & \cellcolor{lightgreen} 31.1 & \underline{17.5} & \cellcolor{lightgreen} 7.4 & 32.3 & \cellcolor{lightgreen} 26.8 & \cellcolor{lightgreen} 15.2 & \underline{5.5} & \cellcolor{lightgreen} 4.3 \\
\bottomrule
\end{tabular}%
}
\end{table*}
 
\begin{table*}[t]
\centering
\caption{\textbf{Alternative Decoding Rule (Dream, block diffusion).} Accuracy (\%) on MBPP and HumanEval under low-confidence decoding as the number of tokens decoded per step increases. Parentheses after each benchmark denote the maximum generation length; column headers denote tokens decoded per step. \colorbox{lightgreen}{Shaded} indicates the best and \underline{underline} indicates the second-best performance in each column.\\}
\label{tab:dream_lowconf_code}

\resizebox{0.75\linewidth}{!}{%
\begin{tabular}{ll cccc ccc}
\toprule
\multirow{2}{*}[-0.3em]{\textbf{Decoding}} & \multirow{2}{*}[-0.4em]{\textbf{\shortstack[l]{Method / \\ Tokens per Step}}} & \multicolumn{4}{c}{\textbf{MBPP (512)}} & \multicolumn{3}{c}{\textbf{HumanEval (512)}} \\
\cmidrule(lr){3-6} \cmidrule(lr){7-9}
 & & \textbf{2} & \textbf{4} & \textbf{8} & \textbf{16} & \textbf{2} & \textbf{4} & \textbf{8} \\
\midrule

\multirow{5}{*}{\textbf{Low Confidence}}
 & Base & 56.0 & 40.1 & 26.8 & 8.6 & 48.8 & 28.7 & 11.6 \\
 & CE-SFT & 52.1 & 33.9 & 15.6 & 2.3 & 37.8 & 31.1 & 14.0 \\
 & CoDD & \underline{62.3} & 48.6 & \underline{33.5} & \underline{8.9} & 50.0 & \underline{37.8} & \cellcolor{lightgreen} 19.5 \\
 & Codpula & \cellcolor{lightgreen} 63.8 & \underline{49.0} & 33.1 & \cellcolor{lightgreen} 13.6 & \underline{50.6} & \cellcolor{lightgreen} 39.6 & 17.1 \\
 & Codpula (Both) & \underline{62.3} & \cellcolor{lightgreen} 51.4 & \cellcolor{lightgreen} 33.9 & \cellcolor{lightgreen} 13.6 & \cellcolor{lightgreen} 54.9 & 33.5 & \underline{17.7} \\
\bottomrule
\end{tabular}%
}
\end{table*}

For the fixed-schedule comparisons among Base, CE-SFT, CoDD, and Codpula, we use the same token-selection rule for each backbone--benchmark pair.
LLaDA-1.0 uses low-confidence decoding on GSM8K, MATH500, and HumanEval, and margin decoding on MBPP. Dream uses low-confidence decoding on the math benchmarks and entropy decoding on the coding benchmarks. LLaDA-2.0 and SDAR use low-confidence decoding. Each rule is held fixed across methods and tokens-per-step settings within the corresponding comparison.

\Cref{tab:llada_decoding_rules} reports additional LLaDA-1.0 results under alternative token-selection rules. Their relative performance varies with the benchmark and degree of parallelism. On GSM8K, Codpula improves over the base model at every evaluated tokens-per-step setting under random decoding. Under margin decoding on the same benchmark, it performs similarly to the base model at 2, 4, and 8 tokens per step, while improving accuracy from $10.1\%$ to $15.7\%$ at 16 tokens per step and from $1.7\%$ to $5.6\%$ at 32 tokens per step. Thus, its gains at high parallelism also extend to alternative token-selection rules.

\Cref{tab:dream_lowconf_code} additionally reports Dream coding results under block diffusion with low-confidence decoding and a maximum generation length of 512 tokens. Both Codpula variants outperform the base model in every reported setting. At four tokens per step, Codpula improves HumanEval accuracy from $28.7\%$ to $39.6\%$, while Codpula (Both) improves MBPP accuracy from $40.1\%$ to $51.4\%$. These results show that Codpula's coding gains also extend to low-confidence decoding.

\subsubsection{Early Stopping Effects}
\label{app:early_stop}

\begin{table*}[h]
\centering
\caption{\textbf{Generation Length and Runtime with Early Stopping.}
Mean statistics on the full AIME24 and AIME25 datasets. All other decoding settings are held fixed: maximum generation length 10{,}240, block length 32, response-anchored block grids, and EOS-based early stopping.}
\label{tab:aime-lowconf-generation}
% \small
\setlength{\tabcolsep}{5pt}
\begin{tabular}{@{}llccc@{}}
\toprule
Dataset & Model &
\shortstack{Tokens\\per step} &
\shortstack{Generated\\tokens} &
\shortstack{Wall time\\(s/q)} \\
\midrule
AIME24 & LLaDA2.0-mini & 2 & 7388.2 & 743.6 \\
       & Codpula      & 2 & 5334.8 & 471.1  \\
\addlinespace
       & LLaDA2.0-mini & 4 & 7870.7 & 411.2 \\
       & Codpula      & 4 & 4080.8 & 163.9 \\
\midrule
AIME25 & LLaDA2.0-mini & 2 & 7137.3 & 710.5 \\
       & Codpula      & 2 & 4358.6 & 359.9  \\
\addlinespace
       & LLaDA2.0-mini & 4 & 8127.8 & 434.5 \\
       & Codpula      & 4 & 6897.7 & 345.4 \\
\bottomrule
\end{tabular}
\end{table*}

\begin{figure}[h]
    \centering
    \includegraphics[width=\linewidth]{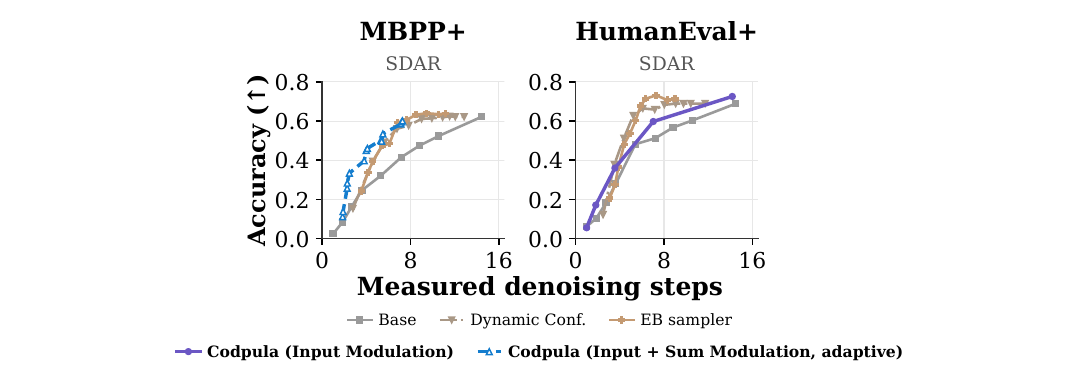}
    \caption{\textbf{SDAR Accuracy versus Denoising Steps per Block.}
    Accuracy on MBPP+ and HumanEval+ as a function of the mean measured number of denoising steps per block. The results use the same evaluation settings as the SDAR panels in \cref{fig:wallclock-llada2-sdar}.
    This view complements the runtime comparison by separating per-block denoising effort from the number of blocks generated before early stopping.}
    \label{fig:sdar_step_appendix}
\end{figure}

We enable EOS-based early stopping for LLaDA-2.0 on AIME and for SDAR on the coding benchmarks. Consequently, fixing the number of tokens decoded per step does not fix the total number of generated tokens or denoising steps: these also depend on when each generation terminates.
\Cref{tab:aime-lowconf-generation} reports mean generation lengths and end-to-end runtimes for LLaDA-2.0-mini and Codpula on AIME24 and AIME25.
At each matched tokens-per-step setting, Codpula produces shorter generations and has lower latency. For example, as shown in table \ref{tab:aime-lowconf-generation}, on AIME24 at four tokens per step, the mean generation length decreases from 7{,}870.7 to 4{,}080.8 tokens, and the mean runtime decreases from 411.2 to 163.9  seconds per question. These latency differences reflect both the cost per denoising step and the amount of generation performed.

Early stopping also explains the nonmonotonic runtime of the SDAR curves in \cref{fig:wallclock-llada2-sdar}. More aggressive parallel decoding can delay EOS prediction, increasing the number of generated blocks enough to offset the reduction in denoising steps per block. For example, the SDAR base model on HumanEval+ executes an average of 31.53 forward passes per question at four tokens per step, compared with 36.09 at eight and 37.45 at sixteen. Thus, increasing parallelism can reduce accuracy while increasing runtime. To separate per-block denoising effort from variation in generation length, \cref{fig:sdar_step_appendix} plots accuracy against the mean number of denoising steps per block. This complements the wall-clock comparison by showing decoding quality at comparable per-block step counts.

\subsubsection{Post-hoc Baseline methods}
\paragraph{Dynamic conf.}
Dynamic confidence-threshold decoding \citep{wu2025fast} adaptively determines the number of tokens unmasked at each decoding step. It accepts predictions at masked positions whose confidence exceeds a fixed threshold, leaving the remaining positions masked for subsequent steps. If no prediction exceeds the threshold, the highest-confidence prediction is accepted to ensure progress. Unlike static decoding, which follows a predetermined token-unmasking schedule, Dynamic Conf.\ allows more tokens to be decoded in parallel when the model is confident, without updating the backbone parameters.

\paragraph{DUS.}
The Dilated Unmasking Scheduler (DUS) \citep{luxembourg2026plan} accelerates masked-diffusion decoding by replacing confidence based token selection with a deterministic, coarse-to-fine schedule. At each denoising step, it reveals a group of positions that are distributed approximately uniformly across the remaining masked sequence; subsequent steps progressively reduce the spacing until the block is complete. By spacing simultaneously decoded tokens apart, dilation reduces their mutual dependence without requiring confidence based ranking. We therefore use the paper's main dilated schedule as the DUS baseline and vary its number of passes per block to compare it with our method at matched computation budgets. With 32 token blocks, the standard DUS schedule supports at most five forward passes. We report the two and four pass settings that match our decoder budgets. Additionally, we disable confidence based remasking because rejected tokens are not regenerated by the fixed schedule, leaving permanent mask holes. For larger budgets in the wall-clock curves, we report \textsc{DUS (ext.)}, which subdivides the final reveal group into consecutive smaller passes while preserving the original position order. For example, a final group of eight positions can be executed as $4+4$ or $2+2+2+2$, with the model reevaluated after each subgroup so later predictions condition on earlier commitments. This extension leaves the preceding dilation schedule unchanged.

\paragraph{ReMDM.}
The remasking diffusion model (ReMDM) sampler \citep{wang2025remasking} augments the reverse diffusion process with self correction: instead of treating every revealed token as permanent, it may remask previously generated tokens and predict them again using the improved surrounding context available at a later denoising step. A time dependent remasking probability $\sigma_t$ controls how aggressively the decoder revises its current solution. The paper studies several schedules for this probability, including capped, rescaled, confidence weighted, switch, and loop based schedules; the switch formulation delays remasking until a chosen stage so that early steps can first construct a coherent draft. We use the switch-style formulation on top of LLaDA's low-confidence decoder because it preserves the original decoding budget while allowing uncertain decisions to be revised.

\paragraph{Tolerator.}
Finish First, Perfect Later (Tolerator) separates generation into a filling stage and a refinement stage \citep{tian2025finish}. The fill up stage first produces a complete draft using the standard diffusion decoding process, with an end-of-text penalty to discourage premature termination. During refinement, a random subset of non-prompt tokens is remasked and regenerated while the remaining draft tokens serve as context for validating these predictions. The remasking rate is gradually reduced with a cosine schedule, enabling broad corrections early in refinement and more stable, localized revisions later. The paper allocates half of the forward-pass budget to each stage. We use this fill and refine strategy as a training-free baseline at matched computation budgets. We use per block refinement because LLaDA2.0's block causal attention is trained with completed preceding blocks; refining each block before proceeding preserves this structure, whereas full draft refinement introduces masks into earlier context.

\subsubsection{Wall-Clock Time Measurement}
\label{app:speed}

All timing measurements use a single NVIDIA RTX PRO 6000 Blackwell GPU with no competing process and a sequence batch size of 1, corresponding to serving one request at a time. GPU isolation is verified before and after each measurement. LoRA adapters are merged into the backbone before timing, and we report median timings over repeated measurements after warm-up. Timing excludes model loading, tokenization, detokenization, evaluation-harness overhead, and grading.

\cref{tab:codpula-overhead} reports Codpula's additional cost per backbone function evaluation (NFE), measured using the same protocol under uniform and adaptive decoding, with and without parameter conditioning.
Measurements use fixed input shapes; adaptive overhead additionally includes a dependence-deferral query at the specified width. Parameter conditioning includes both router evaluation and conditioned circuit computation.

\begin{table}[h]
\centering
\footnotesize
\setlength{\tabcolsep}{6pt}
\caption{\textbf{Codpula decoding overhead on MATH500 (LLaDA-1.0).}
Additional time relative to backbone-only decoding at the same NFE. Measurements use batch size 1, merged LoRA adapters, generation length 512, block length 32, and top-$1{,}024$ evidence candidates. Adaptive results correspond to five Pareto-front operating points;
$dw$ denotes the deferral width.}
\label{tab:codpula-overhead}
\begin{tabular}{@{}llr@{}}
\toprule
Decoding & Operating point & Overhead \\
\midrule
Uniform & Any step budget & $+2.9\%$ \\
\addlinespace
\multirow{5}{*}{Adaptive}
 & NFE $53.8$,  $dw=16$ & $+7.7\%$ \\
 & NFE $74.9$,  $dw=16$ & $+8.2\%$ \\
 & NFE $103.0$, $dw=8$  & $+7.7\%$ \\
 & NFE $131.0$, $dw=5$  & $+6.9\%$ \\
 & NFE $154.2$, $dw=4$  & $+6.8\%$ \\
\bottomrule
\end{tabular}
\end{table}

\subsection{Computational Resources}
All experiments, including training and inference, are conducted on NVIDIA RTX PRO 6000 Blackwell Server Edition GPUs. We observed hardware-dependent differences in experimental results and therefore use the same GPU model throughout to maintain consistent comparisons.

\clearpage
\section{Ablation Studies}
\subsection{Decoding Configurations}

\paragraph{Dream with full-sequence diffusion.}
We also evaluate Dream under full-sequence diffusion with entropy decoding and a generation length of 128 tokens.
\Cref{tab:dream_code_full} shows that both Codpula variants outperform the base model in every reported setting. At four tokens per step, HumanEval accuracy increases from $26.8\%$ for the base model to $32.9\%$ for Codpula and $41.5\%$ for Codpula (Both). Parameter conditioning improves several settings, although Codpula without parameter conditioning achieves higher accuracy on all three benchmarks at eight tokens per step. These results demonstrate that Codpula is also effective under full-sequence diffusion.

\begin{table*}[t]
\centering
\caption{\textbf{Parallel Decoding on Code Benchmarks (Dream, full diffusion).}
Accuracy (\%) under full diffusion with generation length 128 and entropy decoding. Column headers denote tokens decoded per step.
\colorbox{lightgreen}{Shading} and \underline{underlining} indicate the best and second-best accuracy, respectively, in each column.\\}
\label{tab:dream_code_full}
\resizebox{0.8\textwidth}{!}{%
\begin{tabular}{ll ccc ccc ccc}
\toprule
\multirow{2}{*}[-0.3em]{\textbf{Model}} & \textbf{Method} & \multicolumn{3}{c}{\textbf{MBPP}} & \multicolumn{3}{c}{\textbf{MBPP+}} & \multicolumn{3}{c}{\textbf{HumanEval}} \\
\cmidrule(lr){3-5} \cmidrule(lr){6-8} \cmidrule(lr){9-11}
 & {\footnotesize\textit{tokens/step}} & \textbf{2} & \textbf{4} & \textbf{8} & \textbf{2} & \textbf{4} & \textbf{8} & \textbf{2} & \textbf{4} & \textbf{8} \\
\midrule
\multirow{4}{*}{\textbf{Dream}}
 & Base & 50.3 & 42.8 & 32.7 & 46.0 & 38.9 & 31.2 & 34.1 & 26.8 & 24.4 \\
 & CE-SFT & 46.3 & 36.6 & 29.2 & 43.4 & 30.4 & 13.0 & \underline{39.0} & 26.8 & 25.0 \\
 & Codpula & \underline{52.1} & \cellcolor{lightgreen} 49.4 & \cellcolor{lightgreen} 39.3 & \cellcolor{lightgreen} 51.6 & \underline{45.2} & \cellcolor{lightgreen} 35.7 & 34.8 & \underline{32.9} & \cellcolor{lightgreen} 31.7 \\
 & Codpula (Both) & \cellcolor{lightgreen} 55.6 & \underline{48.2} & \underline{38.1} & \underline{49.2} & \cellcolor{lightgreen} 47.9 & \underline{33.3} & \cellcolor{lightgreen} 45.1 & \cellcolor{lightgreen} 41.5 & \underline{29.9} \\
\bottomrule
\end{tabular}%
}
\end{table*}

\paragraph{SDAR with block size 32.}
We additionally evaluate SDAR-8B-Chat-b32 with a matching 32-token circuit, extending the block-size-16 setting used in the main text. \Cref{tab:sdar_block32_results} shows that Codpula improves over the base model at every evaluated setting on MBPP+. On HumanEval+, it improves accuracy at 4--32 tokens per step, matches the base model at two tokens per step, and underperforms at one token per step. At four tokens per step, accuracy increases from 15.9\% to 35.2\% on MBPP+ and from 31.7\% to 40.9\% on HumanEval+. These results show that the benefits of coupling extend to a larger decoding block, particularly when several tokens are committed simultaneously.

\begin{table*}[t]
\centering
\caption{\textbf{Parallel Decoding Performance (SDAR, block size 32).} Accuracy (\%) on HumanEval+ and MBPP+ as the number of tokens decoded per step increases. Column headers denote tokens decoded per step. \colorbox{lightgreen}{Shaded} indicates the best and \underline{underline} indicates the second-best performance in each column.\\}
\label{tab:sdar_block32_results}

\resizebox{\textwidth}{!}{%
\begin{tabular}{ll cccccc cccccc}
\toprule
\multirow{2}{*}[-0.3em]{\textbf{Model}} & \multirow{2}{*}[-0.4em]{\textbf{\shortstack[l]{Method / \\ Tokens per Step}}} & \multicolumn{6}{c}{\textbf{HumanEval+}} & \multicolumn{6}{c}{\textbf{MBPP+}} \\
\cmidrule(lr){3-8} \cmidrule(lr){9-14}
 & & \textbf{1} & \textbf{2} & \textbf{4} & \textbf{8} & \textbf{16} & \textbf{32} & \textbf{1} & \textbf{2} & \textbf{4} & \textbf{8} & \textbf{16} & \textbf{32} \\
\midrule

\multirow{3}{*}{\textbf{SDAR}}
 & Base (static) & \cellcolor{lightgreen} 72.6 & \cellcolor{lightgreen} 59.2 & \underline{31.7} & 12.8 & \underline{5.5} & \underline{3.1} & \underline{59.0} & \underline{39.7} & 15.9 & 3.7 & 0.3 & 0.0 \\
 & CE-SFT & 63.4 & \underline{45.7} & \cellcolor{lightgreen} 40.9 & \underline{15.2} & 4.9 & 0.0 & 53.2 & 34.9 & \underline{20.4} & \underline{5.3} & \underline{0.5} & 0.0 \\
 & Codpula & \underline{68.9} & \cellcolor{lightgreen} 59.2 & \cellcolor{lightgreen} 40.9 & \cellcolor{lightgreen} 18.9 & \cellcolor{lightgreen} 10.4 & \cellcolor{lightgreen} 6.7 & \cellcolor{lightgreen} 63.2 & \cellcolor{lightgreen} 50.3 & \cellcolor{lightgreen} 35.2 & \cellcolor{lightgreen} 19.1 & \cellcolor{lightgreen} 5.6 & \cellcolor{lightgreen} 2.7 \\
\bottomrule
\end{tabular}%
}
\end{table*}

\begin{figure}
    \centering
    \includegraphics[width=0.75\linewidth]{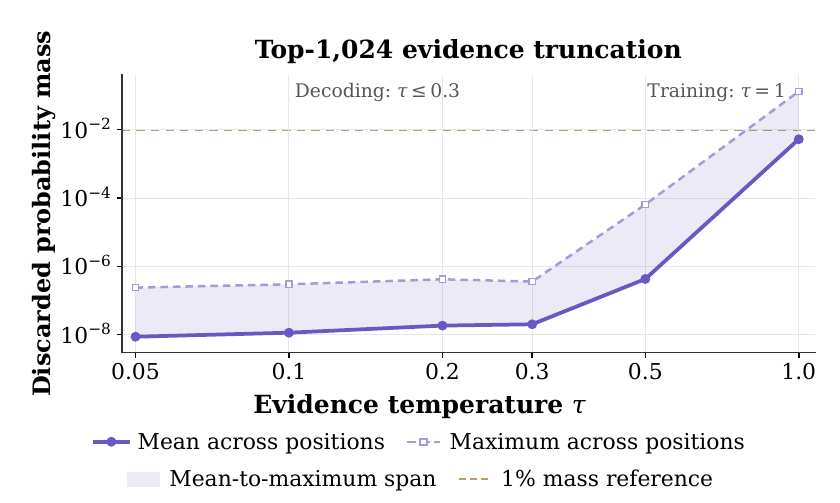}
    \caption{\textbf{Probability mass discarded by top-$1{,}024$ evidence truncation.} Mean and maximum backbone probability mass outside the retained candidates across $46{,}080$ positions on LLaDA-1.0, as a function of evidence temperature $\tau$. Shading spans the mean and maximum; the horizontal dashed line marks $1\%$ mass. At the training temperature $\tau=1$, the discarded mass averages $0.53\%$ and reaches $13.2\%$. At the evaluated decoding temperatures $\tau\leq0.3$, the mean residual is numerically negligible, on the order of $10^{-8}$.}
    \label{fig:evidence_truncation}
\end{figure}

\begin{table}[t]
\centering
\small
\setlength{\tabcolsep}{4pt}
\caption{\textbf{Evidence truncation ablation.}
Accuracy (\%) on GSM8K (200 examples), with $k=1{,}024$. Train and Decode indicate the evidence candidates retained at each stage. The first two rows use the same circuit checkpoint. Results are shown for one checkpoint; all eight matched decoding comparisons yield identical accuracy. \\}
\label{tab:evidence_truncation}
\begin{tabular}{llcccc}
\toprule
\multirow{2}{*}{Train}
& \multirow{2}{*}{Decode}
& \multicolumn{2}{c}{Uniform}
& \multicolumn{2}{c}{Adaptive} \\
\cmidrule(lr){3-4}\cmidrule(lr){5-6}
& & $\tau=0.05$ & $\tau=0.1$
  & $\tau=0.05$ & $\tau=0.1$ \\
\midrule
Top-$k$ & Full    & 63.5 & 68.0 & 65.0 & 64.5 \\
Top-$k$ & Top-$k$ & 63.5 & 68.0 & 65.0 & 64.5 \\
Full    & Full    & 50.5 & 50.5 & 64.0 & 63.0 \\
\bottomrule
\end{tabular}
\end{table}

\subsection{Evidence Truncation}
\label{app:evidence_truncation}
We examine the effect of restricting circuit evidence to the backbone's top-$1{,}024$ candidates per position on LLaDA-1.0. During training, we retain these entries from the backbone's log-probability vector, whereas decoding uses the full vocabulary of $126{,}464$ tokens by default. This truncation restricts the supplied evidence; the circuit's categorical leaves remain parameterized over the full vocabulary.

For backbone logits $\boldsymbol{\ell}$ and evidence temperature $\tau$, we measure the discarded probability mass as
\begin{equation}
    m_k(\boldsymbol{\ell},\tau)
    = 1 - \sum_{v\in S_k(\boldsymbol{\ell})}
    \operatorname{softmax}(\boldsymbol{\ell}/\tau)_v,
    \qquad k=1{,}024,
\end{equation}
where $S_k(\boldsymbol{\ell})$ contains the $k$ highest-scoring tokens. \cref{fig:evidence_truncation} reports measurements over $46{,}080$ positions. At the training temperature $\tau=1$, the mean discarded mass is $0.53\%$, with $19.6\%$ of positions exceeding $1\%$ and a maximum of $13.2\%$. At the evaluated decoding temperatures $\tau\in\{0.05,0.1,0.2,0.3\}$, the measured mean residual is numerically negligible, on the order of $10^{-8}$. Temperature sharpening therefore concentrates nearly all backbone probability mass within the retained candidate set.

We separately ablate truncation during training and decoding on GSM8K with 200 evaluation examples, a maximum generation length of 256, and block length 32. \Cref{tab:evidence_truncation} supports truncating evidence during training, while showing no observed accuracy loss from applying the same truncation at decoding. We use full-vocabulary evidence by default at inference, but top-$1{,}024$ evidence is an empirically equivalent alternative in these comparisons.

\subsection{Contributions of Individual Components}

\begin{figure}[t]
    \centering
    \includegraphics[width=\linewidth]{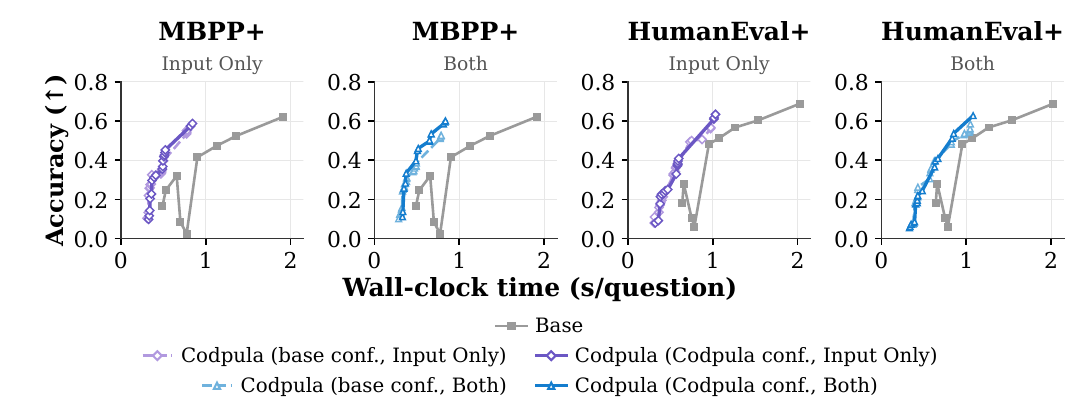}
    \caption{\textbf{Confidence Source for Token Commitment.}
    Accuracy--runtime trade-offs on MBPP+ and HumanEval+ for SDAR with Codpula (Input Only) and Codpula (Both). Both confidence variants use Codpula's token proposals and dependence-based deferral, with candidate positions ranked using either backbone confidence (\emph{base conf.}) or the refined distribution (\emph{Codpula conf.}). The unmodified backbone is included as a reference.}
    \label{fig:sdar_confidence_wall}
\end{figure}

\paragraph{Confidence for token commitment.}
Codpula's refined distributions determine token proposals and can also guide the choice of positions to commit. We ablate this second role on SDAR by comparing confidence scores from the backbone with those from Codpula. Both variants use Codpula's token proposals and the same dependence-based deferral rule; the confidence source determines the ranking of candidate positions. As shown in \cref{fig:sdar_confidence_wall}, Codpula confidence generally improves the accuracy--runtime trade-off at moderate and larger budgets for both Input Only and Both, while backbone confidence remains competitive at the smallest budgets. Thus, the refined distributions provide useful information for the commitment schedule as well as for token prediction.

\paragraph{Stage-1 circuit and backbone adaptation.}

\begin{table*}[ht]
\centering
\caption{\textbf{Circuit and Adapter Ablation on Coding Benchmarks (LLaDA-1.0-8B-Instruct).} Accuracy (\%) for five arms decoded on the same canvas, block grid, and rule; MBPP uses margin and HumanEval uses low-confidence decoding, with $\tau$ swept per arm per budget. \emph{Stage-1 circuit} is our stage-1 circuit decoded without the adapter; \emph{CoDD circuit} is a circuit trained with CoDD's External-additive objective, also without an adapter; \emph{Codpula} is the chosen stage-1 circuit + stage-2 adapter checkpoint. The two difference rows isolate one component each: $\Delta_{\text{adapter}} = \text{Codpula} - \text{Stage-1 circuit}$ holds the circuit fixed and measures what the adapter adds, while $\Delta_{\text{circuit}} = \text{Stage-1 circuit} - \text{CoDD circuit}$ holds the adapter absent and measures circuit quality alone. \colorbox{lightgreen}{Shaded} indicates the best and \underline{underline} indicates the second-best arm in each column.}
\label{tab:coding_ablation}
 
\resizebox{\textwidth}{!}{%
\begin{tabular}{l ccccc ccccc}
\toprule
\textbf{Arm} & \multicolumn{5}{c}{\textbf{MBPP}} & \multicolumn{5}{c}{\textbf{HumanEval}} \\
\cmidrule(lr){2-6} \cmidrule(lr){7-11}
{\footnotesize\textit{tokens/step}} & \textbf{2} & \textbf{4} & \textbf{8} & \textbf{16} & \textbf{32} & \textbf{2} & \textbf{4} & \textbf{8} & \textbf{16} & \textbf{32} \\
\midrule
 
Base & 41.3 & 25.7 & 10.1 & 3.9 & 0.0 & 33.5 & 25.0 & 11.6 & 3.7 & 2.4 \\
CE-SFT (adapter only) & 41.6 & 28.0 & 12.8 & \underline{4.3} & 0.0 & 34.2 & 22.0 & 11.6 & \underline{7.3} & 1.8 \\
Stage-1 circuit (no adapter) & \underline{43.6} & 27.2 & \underline{13.2} & \underline{4.3} & \cellcolor{lightgreen} 1.6 & \cellcolor{lightgreen} 37.8 & \underline{26.8} & \underline{13.4} & \cellcolor{lightgreen} 7.9 & 3.1 \\
CoDD circuit (no adapter) & 43.2 & \underline{29.2} & 9.7 & 2.3 & 0.8 & 34.8 & 25.0 & 12.8 & \cellcolor{lightgreen} 7.9 & \underline{3.7} \\
Codpula (circuit + adapter) & \cellcolor{lightgreen} 48.3 & \cellcolor{lightgreen} 31.9 & \cellcolor{lightgreen} 15.6 & \cellcolor{lightgreen} 5.8 & \underline{1.2} & \underline{37.2} & \cellcolor{lightgreen} 29.3 & \cellcolor{lightgreen} 16.5 & \underline{7.3} & \cellcolor{lightgreen} 4.3 \\
\midrule
 
\quad \footnotesize \textit{$\Delta_{\text{adapter}}$} & \footnotesize\textcolor{posgreen}{+4.7} & \footnotesize\textcolor{posgreen}{+4.7} & \footnotesize\textcolor{posgreen}{+2.3} & \footnotesize\textcolor{posgreen}{+1.6} & \footnotesize $-0.4$ & \footnotesize $-0.6$ & \footnotesize\textcolor{posgreen}{+2.4} & \footnotesize\textcolor{posgreen}{+3.1} & \footnotesize $-0.6$ & \footnotesize\textcolor{posgreen}{+1.2} \\
\quad \footnotesize \textit{$\Delta_{\text{circuit}}$} & \footnotesize\textcolor{posgreen}{+0.4} & \footnotesize $-2.0$ & \footnotesize\textcolor{posgreen}{+3.5} & \footnotesize\textcolor{posgreen}{+2.0} & \footnotesize\textcolor{posgreen}{+0.8} & \footnotesize\textcolor{posgreen}{+3.1} & \footnotesize\textcolor{posgreen}{+1.8} & \footnotesize\textcolor{posgreen}{+0.6} & \footnotesize 0.0 & \footnotesize $-0.6$ \\
\bottomrule
\end{tabular}%
}
\end{table*}

\Cref{tab:coding_ablation} separates the contributions of the stage-1 circuit and the stage-2 backbone adapter on LLaDA-1.0. The stage-1 circuit alone improves over the base model at every reported setting.
Adding the stage-2 adapter yields further gains in seven of ten settings, including 4.7 percentage points on MBPP at both two and four tokens per step and 3.1 points on HumanEval at eight tokens per step.
The complete model also matches or exceeds CE-SFT at every setting. Differences between the stage-1 and CoDD circuits vary in sign, ranging from $-2.0$ to $+3.5$ points. All baselines are independently tuned over learning rates and training budgets, and we report the best-performing configuration for each. The additional gains from stage-2 adaptation, with the stage-1 circuit held fixed, demonstrate the contribution of training the backbone through the circuit.

\begin{wrapfigure}{r}{0.4\textwidth}
    \centering
    \includegraphics[width=\linewidth]{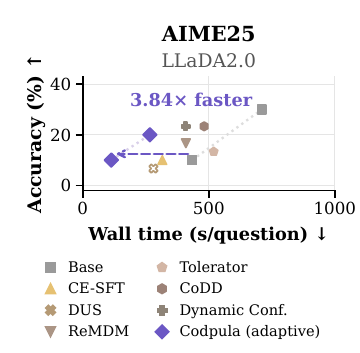}
    \caption{\textbf{Additional AIME25 results with LLaDA-2.0-mini.}
    Accuracy (\%) versus wall-clock time per question.
    Adaptive Codpula achieves a $3.84\times$ speedup over the
    base model at matched $10\%$ accuracy.}
    \label{fig:aime25_additional}
\end{wrapfigure}

\subsection{Circuit Removal at Inference}
\label{app:circuit_removal}

Removing the PC at inference is not an appropriate way to isolate its contribution. The adapted backbone and PC together define Codpula's normalized conditional distribution, and the adapter is trained under this complete distribution's conditional-likelihood objective. The backbone outputs serve as token potentials within the circuit and are not optimized as standalone denoising probabilities. Removing the PC and decoding directly from these potentials therefore substitutes a different distribution for the trained model. We consequently do not report circuit-removal results as a component ablation. Backbone adaptation without a circuit is instead evaluated using the independently trained CE-SFT baseline.

\subsection{Evidence-Temperature Sensitivity}
\label{app:temperature_sensitivity}

We vary the evidence temperature $\tau$ while holding the LLaDA-1.0 backbone adapter, circuit, and decoding rule fixed. \Cref{tab:temperature_sensitivity} shows that the preferred temperature depends on the decoding budget. Among the temperatures shown, the best value shifts from $\tau=0.05$ at 256 steps to $\tau=0.5$ at 32 steps, with $\tau=0.1$ preferred at the intermediate budgets. A finer sweep at 32 steps gives $11.2\%$, $11.6\%$, and $11.2\%$ accuracy at $\tau=0.5$, $0.6$, and $0.7$, respectively. These results motivate selecting the evidence temperature separately for each decoding budget.

\begin{table}[t]
\centering
\small
\setlength{\tabcolsep}{6pt}
\caption{\textbf{Evidence-temperature sensitivity on MATH500.}
Accuracy (\%) with LLaDA-1.0 under uniform low-confidence decoding, using a fixed adapter and circuit, generation length 512, and block length 32. The table reports the temperature grid shared by all four budgets. Bold indicates the best accuracy among the displayed
temperatures in each row.}
\label{tab:temperature_sensitivity}
\begin{tabular}{rrrrrr}
\toprule
Tokens/step & Steps & $\tau=0.05$ & $\tau=0.1$ & $\tau=0.3$ & $\tau=0.5$ \\
\midrule
2  & 256 & \textbf{39.6} & 36.2 & 39.4 & 37.6 \\
4  & 128 & 29.4 & \textbf{34.8} & 31.8 & 29.4 \\
8  & 64  & 17.8 & \textbf{19.4} & 15.4 & 17.4 \\
16 & 32  & 6.8  & 10.0 & 10.8 & \textbf{11.2} \\
\bottomrule
\end{tabular}
\end{table}
 
\subsection{Additional Results}

\subsubsection{AIME Benchmarks}
\label{app:additional_aime}

\begin{figure}[t]
    \centering
    \includegraphics[width=\linewidth]{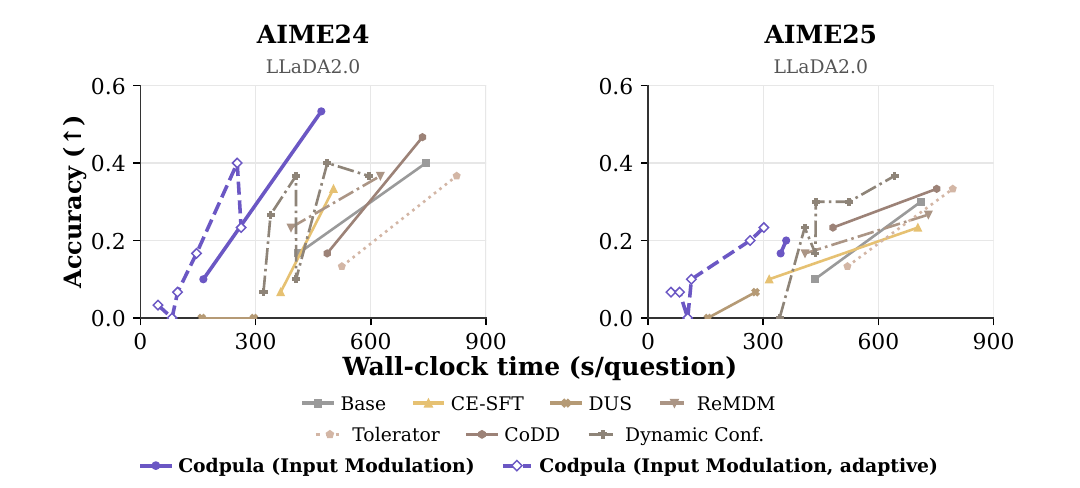}
    \caption{\textbf{Accuracy--time trade-offs on AIME24 and AIME25.}
    Results use LLaDA-2.0-mini. Codpula (Input Only) is evaluated with uniform decoding (solid purple) and dependence-adaptive decoding (dashed purple). Each marker denotes an evaluated decoding configuration; lines connect configurations of the same method.}
    \label{fig:aime_additional_curves}
\end{figure}

We provide additional accuracy--time comparisons on AIME24 and AIME25 using LLaDA-2.0-mini. On AIME25, adaptive Codpula matches the base model's $10\%$ accuracy with a $3.84\times$ speedup (\cref{fig:aime25_additional}). \Cref{fig:aime_additional_curves} shows results across decoding configurations on both benchmarks. On AIME24, uniform Codpula reaches $53.3\%$ accuracy, while adaptive Codpula matches the base model's $40\%$ accuracy at a lower time cost. On AIME25, the gains are concentrated in the lower-latency region; several baselines attain higher accuracy at larger time budgets.

\clearpage
\section{Generated Examples}
\begin{figure}[h]
    \centering
    \fbox{\begin{minipage}{0.95\linewidth}
        \textbf{Baseline (LLaDA) -- 32 steps, 8 tokens/step} \\ \hrule \vspace{0.2cm}
        \textbf{Question:} There are 9,300 pennies in a cup. What is the total dollar amount in a
        stack that contains two thirds of the pennies in the cup? \\
        \textbf{Generation:} \\
        Step 1: Calculate the number of pennies in the stack. \dots\ we multiply 9,300 by
        $\frac{2}{3}$:
        \[ \text{Number of} \times \tfrac{2}{3} = 9{,}300 \times \tfrac{2}{3} = \textcolor{red}{6{,}300} \]
        Step 2: \dots\ we multiply the of of pennies the stack by \$0.01:
        \[ \text{Total dollar} = 6{,}300 \times 0.01 = 63 \]
        Therefore, the total dollar amount in the\{ is $\boxed{63}$.
        \vspace{0.2cm} \\
        \textbf{Status:} \textcolor{red}{\textbf{Incorrect (one digit of a parallel-committed number)}}
    \end{minipage}}

    \vspace{0.3cm}

    \fbox{\begin{minipage}{0.95\linewidth}
        \textbf{CoDD -- 32 steps, 8 tokens/step} \\ \hrule \vspace{0.2cm}
        \textbf{Generation:} \\
        \[ \tfrac{2}{3 \times 9300} = 1800000000 = \textcolor{posgreen}{6200} \]
        So, the total number of pennies in the stack is: $\boxed{6200}$ \\
        Step 3: \dots\ Each penny is worth \$0.01.
        \[ \text{Total dollar amount} = 6200 \times 0.01 = \textcolor{red}{6} \]
        Therefore, the total amount of dollars in the stack is: $\boxed{6}$
        \vspace{0.2cm} \\
        \textbf{Status:} \textcolor{red}{\textbf{Incorrect (correct count, digit lost at the unit conversion)}}
    \end{minipage}}

    \vspace{0.3cm}

    \fbox{\begin{minipage}{0.95\linewidth}
        \textbf{Codpula (Ours) -- 32 steps, 8 tokens/step} \\ \hrule \vspace{0.2cm}
        \textbf{Generation:} \\
        Step 1: \dots\ the stack contains two thirds of these pennies. Therefore:
        \[ \text{Number of pennies in the stack} = \tfrac{2}{3} \times 9300 = \textcolor{posgreen}{6200} \]
        Step 2: Since each penny is worth \$0.01, the total dollar amount in the stack is:
        \[ \text{Total in dollars in the stack} = 6200 \times 0.01 = \textcolor{posgreen}{62} \]
        Therefore, the total dollar amount in the stack is $\boxed{62}$.
        \vspace{0.2cm} \\
        \textbf{Status:} \textcolor{posgreen}{\textbf{Correct}}
    \end{minipage}}
    \caption{\textbf{Multi-digit arithmetic at high parallelism (GSM8K, 32 steps of 8 tokens).}
    Both failures are single-digit corruptions of a quantity emitted inside one parallel commit
    window, not reasoning errors: the baseline writes $6{,}300$ for $9{,}300 \times \frac{2}{3}$,
    and CoDD recovers $6200$ but then drops a digit converting to dollars. Codpula carries both
    quantities intact. CE-SFT (not shown) degenerates into a run of zeros at this budget.}
    \label{fig:qual_gsm8k_digits}
\end{figure}

\begin{figure}[h]
    \centering
    \fbox{\begin{minipage}{0.95\linewidth}
        \textbf{Baseline (LLaDA) -- 64 steps} \\ \hrule \vspace{0.2cm}
        \textbf{Question:} Simplify $(u+4)(u-1) - (u-3)(u+6)$. \\
        \textbf{Generation:} \\
        1. \textbf{Expand each bin of parentheses:}
        \[ (u + 4)(u - 1) = \textcolor{red}{u^2 - 4u + 4} \qquad (u-3)(u+6) = u^2 + 3u - 18 \]
        3. \textbf{Distribute the negative sign:} \[ = u^2 - 4u + 4 - u^2 - 3u + 18 \]
        4. \textbf{Combine like terms:} \[ = (u^2 - u^2) + (-4u - 3u) + (4 + 18) \]
        5. \[ = 0 \] \dots\ $\boxed{22}$ \emph{(box left unclosed)}
        \vspace{0.2cm} \\
        \textbf{Status:} \textcolor{red}{\textbf{Incorrect (wrong expansion, then the $u$ term is dropped)}}
    \end{minipage}}

    \vspace{0.3cm}

    \fbox{\begin{minipage}{0.95\linewidth}
        \textbf{Codpula (Ours) -- 64 steps} \\ \hrule \vspace{0.2cm}
        \textbf{Generation:} \\
        \textbf{Step 1: Expand each binomial}
        \[ (u+4)(u-1) = u^2 - u + 4u - 4 = \textcolor{posgreen}{u^2 + 3u - 4} \]
        \[ (u-3)(u+6) = u^2 + 6u - 3u - 18 = \textcolor{posgreen}{u^2 + 3u - 18} \]
        \textbf{Step 2: Subtract the second expansion from the first}
        \[ (u^2 + 3u - 4) - (u^2 + 3u - 18) = \textcolor{posgreen}{14} \]
        \[ \boxed{14} \]
        \vspace{0.2cm} \\
        \textbf{Status:} \textcolor{posgreen}{\textbf{Correct}}
    \end{minipage}}
    \caption{\textbf{Algebraic simplification at low compute (MATH500, 64 steps).} The baseline
    mis-expands the first binomial ($-4u$ for $+3u$) and then, combining, keeps only the constant
    pair $(4+18)$ and discards the $u$ terms it had just written. Codpula keeps both expansions
    consistent and cancels correctly. CE-SFT (not shown) collapses into a repeated token at this
    budget.}
    \label{fig:qual_math500_algebra}
\end{figure}

% ------------------------------- adaptive deferral traces -------------------------------

\begin{figure}[h]
    \centering
    \fbox{\begin{minipage}{0.95\linewidth}
        \textbf{Deferral trace -- HumanEval \texttt{mean\_absolute\_deviation}, $\delta=0.05$, width 16} \\
        \hrule \vspace{0.2cm}
        \textbf{Step $t$: block empty (0/32 committed).} The confidence stage proposes 16 slots;
        the circuit scores each and the decoder commits $|\mathrm{dep}|<\delta$.

        \vspace{0.15cm}
        {\small\begin{tabular}{@{}llr l@{}}
        \toprule
        pos & token & $|\mathrm{dep}|$ & decision \\
        \midrule
        14 & \texttt{' numbers'} & $0.7426$ & \textcolor{red}{\textbf{defer}} \\
        15 & \texttt{':'}        & $0.7418$ & \textcolor{red}{\textbf{defer}} \\
        18 & \texttt{' return'}  & $0.3985$ & \textcolor{red}{\textbf{defer}} \\
        21 & \texttt{'\ \ \ '}   & $0.0727$ & \textcolor{red}{\textbf{defer}} \\
        \midrule
        11 & \texttt{'\ \ \ '}   & $0.0022$ & commit \\
         6 & \texttt{'0'}        & $0.0006$ & commit \\
         7 & \texttt{'\textbackslash n'} & $0.0003$ & commit \\
         4 & \texttt{'1'}        & $0.0000$ & commit \\
         5 & \texttt{'.'}        & $0.0000$ & commit \\
        \multicolumn{4}{@{}l}{\quad\emph{\dots\ 7 further tokens, all $|\mathrm{dep}|\le0.0022$, all committed}} \\
        \bottomrule
        \end{tabular}}

        \vspace{0.25cm}
        \textbf{Step $t{+}1$: 12/32 now committed.} The deferred tokens are re-scored against the
        context that just landed:

        \vspace{0.15cm}
        {\small\begin{tabular}{@{}llrrl@{}}
        \toprule
        pos & token & $|\mathrm{dep}|$ at $t$ & at $t{+}1$ & decision \\
        \midrule
        18 & \texttt{' return'}  & $0.3985$ & $\mathbf{0.0004}$ & \textcolor{posgreen}{commit} \\
        14 & \texttt{' numbers'} & $0.7426$ & $\mathbf{0.0208}$ & \textcolor{posgreen}{commit} \\
        15 & \texttt{':'}        & $0.7418$ & $\mathbf{0.0209}$ & \textcolor{posgreen}{commit} \\
        \bottomrule
        \end{tabular}}
    \end{minipage}}
    \caption{\textbf{What the deferral gate holds back, and why it lets go.} At step $t$ the gate
    defers exactly 4 of 16 proposed slots, and those four are precisely the tokens of the guard
    clause \texttt{if not numbers: return \dots} mutually determined, and wrong in a way that
    compounds if any one is committed against a different choice of the others. The twelve it
    commits are indentation, newlines and list markers, scoring $|\mathrm{dep}|\le 0.0022$ and most
    of them exactly $0$. One step later the surrounding context has been written and the same three
    tokens score $20$--$1000\times$ lower, so the gate releases them. Deferral is therefore not a
    rejection: it is a re-ordering that lets an entangled group be decided jointly rather than
    independently.}
    \label{fig:qual_defer_return}
\end{figure}

\begin{figure}[h]
    \centering
    \fbox{\begin{minipage}{0.95\linewidth}
        \textbf{Deferral trace -- HumanEval \texttt{string\_xor}, $\delta=0.05$, width 16} \\
        \hrule \vspace{0.2cm}
        \textbf{Step $t$: 16/32 committed.} The model is emitting the docstring example
        \texttt{>>> string\_xor('010', '110')}:

        \vspace{0.15cm}
        {\small\begin{tabular}{@{}llrl@{}}
        \toprule
        pos & token & $|\mathrm{dep}|$ & decision \\
        \midrule
        25 & \texttt{' >>>'}     & $1.7571$ & \textcolor{red}{\textbf{defer}} \\
        27 & \texttt{'\_x'}      & $1.2851$ & \textcolor{red}{\textbf{defer}} \\
        26 & \texttt{' string'}  & $1.1281$ & \textcolor{red}{\textbf{defer}} \\
        29 & \texttt{"('"}       & $0.2425$ & \textcolor{red}{\textbf{defer}} \\
        28 & \texttt{'or'}       & $0.2262$ & \textcolor{red}{\textbf{defer}} \\
        31 & \texttt{'1'}        & $0.1353$ & \textcolor{red}{\textbf{defer}} \\
        30 & \texttt{'0'}        & $0.0845$ & \textcolor{red}{\textbf{defer}} \\
        16 & \texttt{' return'}  & $0.0734$ & \textcolor{red}{\textbf{defer}} \\
        \midrule
        20 & \texttt{' a'}       & $0.0033$ & commit \\
        24 & \texttt{'\ \ \ '}   & $0.0002$ & commit \\
        21 & \texttt{' string'}  & $0.0000$ & commit \\
        22 & \texttt{'.'}        & $0.0000$ & commit \\
        23 & \texttt{'\textbackslash n'} & $0.0000$ & commit \\
        \bottomrule
        \end{tabular}}

        \vspace{0.25cm}
        \textbf{Step $t{+}1$: 21/32 committed.}
        The dependence scores change as follows:
        
        \vspace{0.15cm}
        {\small
        \begin{tabular}{@{}lrrl@{}}
        \toprule
        token & $|\mathrm{dep}|$ at $t$ & at $t{+}1$ & decision \\
        \midrule
        \texttt{' return'} & $0.0734$ & $\mathbf{0.0150}$
            & \textcolor{posgreen}{commit} \\
        \texttt{'or'} & $0.2262$ & $\mathbf{0.0354}$
            & \textcolor{posgreen}{commit} \\
        \texttt{' as'} & $0.0761$ & $\mathbf{0.0008}$
            & \textcolor{posgreen}{commit} \\
        \midrule
        \texttt{' >>>'} & $1.7571$ & $>\delta$
            & \textcolor{red}{\textbf{defer}} \\
        \texttt{'\_x'} & $1.2851$ & $>\delta$
            & \textcolor{red}{\textbf{defer}} \\
        \texttt{' string'} & $1.1281$ & $>\delta$
            & \textcolor{red}{\textbf{defer}} \\
        \bottomrule
        \end{tabular}}
    \end{minipage}}
    \caption{\textbf{The deferral trace tracks name agreement.} The deferred group spells
    the identifier \texttt{string\_xor} and its call literal: \texttt{' >>>'}, \texttt{' string'},
    \texttt{'\_x'}, \texttt{'or'}, \texttt{"('"}, \texttt{'0'}, \texttt{'1'}. These must agree with
    the function actually defined above, so committing them independently is exactly how a parallel
    decode invents a name that does not exist. Tokens with no such constraint (the word
    \texttt{' string'} at position 21, a period, a newline, indentation) score identically
    $0.0000$ and are committed in the same step. Note the two \texttt{' string'} occurrences:
    position 26, part of the identifier, scores $1.1281$; position 21, ordinary prose, scores $0$.
    The signal is contextual entanglement, not the token's identity.}
    \label{fig:qual_defer_identifier}
\end{figure}

\clearpage

\section{Algorithm Details}
\subsection{Dependence Queries and Decoding Cost}
\label{app:decoding-details}

We evaluate \cref{eq:dep-score} using marginal queries to a single
circuit instantiated from the current neural outputs
$\params_t=f_{\vparams}(\x_t)$.
The neural token potentials, coupling gates, and local normalizers
remain fixed across these queries; only the observed assignments change.
All probabilities below refer to the circuit used for dependence scoring,
with its evidence temperature and vocabulary support held fixed.

\paragraph{Computing the dependence scores.}
For $S\subseteq\mathcal C$, let $e_S$ retain the currently visible
tokens, additionally fix positions in $S$ to their proposed values
$\hat\x^S$, and marginalize all remaining positions.
Write
\begin{equation}
    \ell(S)
    :=\log\tilde p(e_S;\params_t,\wparams).
    \label{eq:app-decode-log-evidence}
\end{equation}
In particular, $e_{\varnothing}$ contains only the currently visible
tokens, so $\exp\ell(\varnothing)$ is the shared visible-evidence
normalizer. Assuming the queried assignments have positive probability,
\begin{equation}
\begin{aligned}
    \log p_{\vparams,\wparams}(\hat\x^S\mid\x_t)
        &=\ell(S)-\ell(\varnothing),\\
    \log p_{\vparams,\wparams}
        (\hat x^i\mid\hat\x^{\mathcal C\setminus\{i\}},\x_t)
        &=\ell(\mathcal C)
          -\ell(\mathcal C\setminus\{i\}).
\end{aligned}
    \label{eq:app-decode-conditionals}
\end{equation}
Substituting into \cref{eq:dep-score} gives
\begin{equation}
    \delta_i
    =\left|
        \ell(\mathcal C)
        -\ell(\mathcal C\setminus\{i\})
        -\ell(\{i\})
        +\ell(\varnothing)
      \right|.
    \label{eq:app-decode-batched-score}
\end{equation}
Thus, for $m=|\mathcal C|$, the scores use one complete-candidate
marginal, $m$ singleton marginals, and $m$ leave-one-out marginals.
Computing these $2m+1$ probabilities requires at most $2m+2$ circuit
evaluations, including the shared normalizer. A cached normalizer can
be reused only when it was computed with the same circuit parameters,
neural evidence, and visible assignments.
Duplicate queries can also be reused; for $m=1$, the score is zero
without any additional query.

\paragraph{Batched circuit evaluation.}
We batch the evidence assignments
$e_{\varnothing}$, $e_{\mathcal C}$,
$\{e_{\{i\}}:i\in\mathcal C\}$, and
$\{e_{\mathcal C\setminus\{i\}}:i\in\mathcal C\}$
in one circuit call. Each batch item uses the same modulated circuit
and differs only in which variables are observed.
At an input node, an observed variable contributes its probability
under the modulated input distribution, while an unobserved variable
contributes the local normalizer
\(\log Z_{n,i}\). The bottom-up computation in
Appendix~\ref{app:pc-normalization} then returns $\ell(S)$ for each
evidence assignment, marginalizing over all unresolved tokens and
circuit latent variables.
The input normalizers can be shared across the batch.
After input preparation, the total circuit work is $O(m|E|)$ for
$m\geq2$, where $|E|$ is the number of circuit edges.
Batching parallelizes these evaluations and requires no additional
Transformer evaluation.

\paragraph{Computation and measured overhead.}
Dependence evaluation adds circuit computation but reuses
the current neural outputs, requiring no additional
backbone evaluation within a step.
Figures~1 and~2 report the resulting accuracy--latency
trade-offs under adaptive decoding.
Table~\ref{tab:codpula-overhead} quantifies the additional
cost relative to backbone-only decoding at matched NFE,
while Appendix~\ref{app:speed} describes the
measurement protocol.

\begin{algorithm}[ht]
\caption{Multiplicative update from signed circuit flows}
\label{alg:signed-flow-update}
\begin{algorithmic}[1]
\Require Normalized positive weights $\wparams$; batch flows $F^+,F^-\geq0$;
         step size $\eta\geq0$
\State Compute $\operatorname{TD}_{\wparams}(n)$ for all sum nodes $n$
\ForAll{sum nodes $n$}
  \If{$\operatorname{TD}_{\wparams}(n)=0$}
    \State $\wparams^+_{n,c}\gets\wparams_{n,c}$ for all $c\in\ch(n)$
  \Else
    \State $T_{\wparams}(n,c)\gets
       \operatorname{TD}_{\wparams}(n)\wparams_{n,c}$ for all $c\in\ch(n)$
    \State $R_{n,c}\gets
       \dfrac{T_{\wparams}(n,c)+\eta F^+(n,c)}
             {T_{\wparams}(n,c)+\eta F^-(n,c)}$ for all $c\in\ch(n)$
    \State $\wparams^+_{n,c}\gets
       \dfrac{\wparams_{n,c}R_{n,c}}
             {\sum_{d\in\ch(n)}\wparams_{n,d}R_{n,d}}$
       for all $c\in\ch(n)$
  \EndIf
\EndFor
\State \Return $\wparams^+$
\end{algorithmic}
\end{algorithm}

\begin{algorithm}[ht]
\caption{Joint learning from a minibatch of denoising examples}
\label{alg:joint-learning}
\begin{algorithmic}[1]
\Require Minibatch $\mathrm B$; parameters $\wparams,\vparams$;
         circuit step size $\eta\geq0$
\State $F^+,F^-\gets0$;\quad $\boldsymbol{g}_{\vparams}\gets\boldsymbol{0}$
\ForAll{$(t,\x_0,\x_t)\in\mathrm B$}
  \State $\params\gets f_{\vparams}(\x_t)$;\quad
         $a\gets w(t)/|\mathrm B|$
  \State Construct $e^+$ and $e^-$
  \State $(\ell^\pm,F_{e^\pm},\boldsymbol{g}^\pm)
         \gets\operatorname{CircuitPass}(\wparams,\params;e^\pm)$
         \Comment{one pass for each evidence assignment}
  \State $F^\pm\gets F^\pm+aF_{e^\pm}$
  \State $\boldsymbol{g}_{\vparams}\gets\boldsymbol{g}_{\vparams}
         +a\,\mathbf{J}_{\vparams}(\x_t)^\top
          (\boldsymbol{g}^+-\boldsymbol{g}^-)$
\EndFor
\State $\wparams^+\gets
       \operatorname{SignedFlowUpdate}(\wparams;F^+,F^-,\eta)$
       \Comment{Algorithm~\ref{alg:signed-flow-update}}
\State $\vparams^+\gets
       \operatorname{Opt}(\vparams;\boldsymbol{g}_{\vparams})$
       \Comment{gradient ascent on $\calL_{\mathrm B}$}
\State \Return $\wparams^+,\vparams^+$
\end{algorithmic}
\end{algorithm}

\begin{algorithm}[t]
\caption{Dependence-adaptive decoding of one block}
\label{alg:decode}
\begin{algorithmic}[1]
\Require Current sequence $x$; active block $\mathcal I$;
candidate width $w$; threshold $\tau>0$; step cap $M$
\For{$k=1,\ldots,M$}
    \State $\mathcal U\gets
        \{i\in\mathcal I:x^i=\texttt{[MASK]}\}$
    \If{$\mathcal U=\emptyset$}
        \State \textbf{break}
    \EndIf
    \State Compute proposals $\hat x$, confidences $c$,
        and scoring distribution $q_t^{\mathrm{dep}}$
        from the current sequence $x$
        \Comment{one backbone evaluation}
    \If{$k=M$}
        \State $\mathcal S\gets\mathcal U$
            \Comment{complete the block}
    \Else
        \State Select a set of candidates $\mathcal C$
        \If{$|\mathcal C|=1$}
            \State $\mathcal S\gets\mathcal C$
        \Else
            \State Compute $\delta_i$ for
                $i\in\mathcal C$ using
                \cref{eq:dep-score}
            \State $\mathcal S\gets
                \{i\in\mathcal C:\delta_i<\tau\}$
            \If{$\mathcal S=\emptyset$}
                \State $\mathcal S\gets
                    \{\arg\min_{i\in\mathcal C}\delta_i\}$
            \EndIf
        \EndIf
    \EndIf
    \State $x^i\gets\hat x^i$ for all $i\in\mathcal S$
\EndFor
\State \Return $x$
\end{algorithmic}
\end{algorithm}

\clearpage
\section{Derivation of the Joint Denoising Objective}
\label{app:elbo-derivation}

We derive \cref{eq:codd-objective} as a lower-bound
training objective obtained from the diffusion ELBO.
Consider sequences of fixed length $L$ under independent
absorbing masking, where $\alpha_t$ denotes the masking
probability. Assume a differentiable increasing schedule
with $\alpha_0=0$ and $\alpha_1=1$.
The joint denoiser
$p_{\vparams,\wparams}(\x_0\mid\x_t)$ assigns probability
only to clean completions consistent with the visible
tokens.

The continuous-time reverse process reveals each masked
position at rate $\dot\alpha_t/\alpha_t$, using the
corresponding token marginal of the joint denoiser.
Writing $\mathcal M_t$ for the masked positions, the
standard diffusion ELBO takes the form
~\citep{sahoo2024simple,shi2024simplified}
\begin{equation}
    \mathcal E(\wparams,\vparams)
    =
    \int_0^1
        \frac{\dot\alpha_t}{\alpha_t}
        \mathbb E_{\x_0,\x_t}\!\left[
            \sum_{i\in\mathcal M_t}
            \log p_{\vparams,\wparams}(x_0^i\mid\x_t)
        \right]\,\mathrm dt,
    \label{eq:app-joint-diffusion-elbo}
\end{equation}
where $\x_0\sim p_{\mathrm{data}}$ and
$\x_t\sim q_t(\cdot\mid\x_0)$.
Here, each token probability is a marginal of the joint
denoiser; no factorization of its clean-completion
distribution is assumed.
We assume the usual integrability and endpoint conditions
for the continuous-time bound.

To relate this expression to the conditional joint
likelihood, observe that each token marginal sums over
completions that include $\x_0$. Consequently,
\begin{equation}
    p_{\vparams,\wparams}(x_0^i\mid\x_t)
    \geq
    p_{\vparams,\wparams}(\x_0\mid\x_t).
    \label{eq:app-joint-marginal-bound}
\end{equation}
Since $|\mathcal M_t|\leq L$ and $\log p_{\vparams, \wparams} (x_0 | x_t) \leq 0$
, it follows that
\begin{equation}
\begin{aligned}
    \sum_{i\in\mathcal M_t}
        \log p_{\vparams,\wparams}(x_0^i\mid\x_t)
    &\geq
        |\mathcal M_t|\,
        \log p_{\vparams,\wparams}(\x_0\mid\x_t)\\
    &\geq
        L\log p_{\vparams,\wparams}(\x_0\mid\x_t).
\end{aligned}
    \label{eq:app-joint-likelihood-bound}
\end{equation}
Since $\dot\alpha_t/\alpha_t\geq0$ for $t\in(0,1)$, taking expectations and integrating \cref{eq:app-joint-likelihood-bound} with this nonnegative weight preserves the inequality. Substituting into
\cref{eq:app-joint-diffusion-elbo} therefore gives
\begin{equation}
    \mathcal E(\wparams,\vparams)
    \geq
    L\int_0^1
        \frac{\dot\alpha_t}{\alpha_t}
        \mathbb E_{\x_0,\x_t}\!\left[
            \log p_{\vparams,\wparams}(\x_0\mid\x_t)
        \right]\,\mathrm dt.
    \label{eq:app-joint-elbo-bound}
\end{equation}

For timesteps sampled from a density $\rho(t)>0$,
importance weighting expresses the integral as
\begin{equation}
\begin{aligned}
    \calL(\wparams,\vparams)
    &=
    \mathbb E_{t\sim\rho,\x_0,\x_t}\!\left[
        w(t)\log p_{\vparams,\wparams}(\x_0\mid\x_t)
    \right],\\
    w(t)&=\frac{\dot\alpha_t}{\alpha_t\rho(t)}.
\end{aligned}
    \label{eq:app-joint-objective}
\end{equation}
Thus, $L\calL(\wparams,\vparams)\leq
\mathcal E(\wparams,\vparams)$.
Because $L$ is fixed and positive, maximizing this
lower bound is equivalent to maximizing
\cref{eq:codd-objective}.
For uniform timestep sampling and $\alpha_t=t$,
the weight is $w(t)=1/t$.
\clearpage

\section{Additional Details on Probabilistic Circuits}

\subsection{Structure and Exact Inference}
\label{app:pc-normalization}
We state the structural properties that guarantee tractable computation of $Z$. Let $\scope(n)$ denote the variables appearing in the input nodes descending from node $n$.

\begin{definition}[Smoothness]
A PC is smooth if all children of every sum node $n$
have the same scope:
\[
    \scope(c)=\scope(n),
    \qquad \forall c\in\ch(n).
\]
\end{definition}

\begin{definition}[Decomposability]
A PC is decomposable if the children of every product
node $n$ have pairwise disjoint scopes:
\[
    \scope(c)\cap\scope(d)=\varnothing,
    \qquad \forall c,d\in\ch(n),\ c\neq d.
\]
\end{definition}

\paragraph{Bottom-up evaluation.}
Given observed token assignments $e$, let $L_n(e)$ denote the probability of the evidence within $\scope(n)$. At an input node, we evaluate the observed token's probability or sum over all token values if the variable is unobserved. Product nodes combine evidence from disjoint scopes by multiplication, while sum nodes average their children's evidence probabilities using the mixture weights:
\[
    L_n(e)=
    \begin{cases}
        g_n(e_i),
            & n\text{ is an input over observed }X_i,\\
        \sum_{v\in\mathcal V}g_n(v)=1,
            & n\text{ is an input over unobserved }X_i,\\
        \prod_{c\in\ch(n)}L_c(e),
            & n\text{ is a product node},\\
        \sum_{c\in\ch(n)}\omega_{n,c}L_c(e),
            & n\text{ is a sum node}.
    \end{cases}
\]
Smoothness and decomposability ensure that these local operations compute the exact evidence probability. Evaluating each node after its children yields this probability at the root in a single pass, with cost linear in the number of circuit edges after input
evaluation.

\paragraph{Mapping the modulated circuit to tractable inference.}
Appendix~D of \citet{li2026breaking} establishes tractable
normalization by expressing its normalizer as an
independent soft-evidence query on a PC.
We follow the same reasoning, accounting for modulation
of both input distributions and sum weights.

Fix the masked context $\x_t$ and neural outputs
$\params_t=f_{\vparams}(\x_t)$.
The input modulation in \cref{eq:token-potentials}
preserves each input node's univariate scope and produces
a normalized distribution.
Likewise, the gates in \cref{eq:conditioned-edge} produce
nonnegative, normalized sum weights without changing
the children of any node.
Assuming positive local normalizers, the resulting
$\tilde p(\x_0;\params_t,\wparams)$ is therefore a
normalized PC with the same smoothness and
decomposability properties as the original circuit.
Although the neural outputs depend on the full context,
they remain fixed as we sum over $\x_0$.

The normalizer in \cref{eq:Z} can consequently be written as
\begin{equation}
    Z(\params_t,\wparams)
    =
    \sum_{\x_0}
        \tilde p(\x_0;\params_t,\wparams)
        \prod_{i:\,x_t^i\neq\texttt{[MASK]}}
            \mathbf 1\{x_0^i=x_t^i\}.
    \label{eq:app-modulated-normalizer}
\end{equation}
The indicators are univariate evidence functions, so
\cref{eq:app-modulated-normalizer} is an independent-evidence
query on the modulated PC.
The tractable inference result of
\citet[Theorem~1]{liu2024image} therefore applies.

Concretely, each observed input is evaluated at its visible token, while each masked input sums to one. These values propagate through the product nodes and the sum nodes using their gated weights, yielding $Z$ at the root. Once the modulated parameters and input values have been computed, this bottom-up pass is exact and linear in the number of circuit edges.

\subsection{Correspondence with Standard Circuit Learning Algorithms}
\label{app:anemone-background}

We make explicit how \cref{eq:anemone} recovers the basic
updates of full-batch EM, interpolated mini-batch EM, and
Anemone. Consider a standalone smooth, decomposable PC
with normalized inputs and strictly positive, normalized
sum weights. For a batch $\mathrm B$, define
\begin{equation}
    F_{\mathrm B}^{+}(n,c)
    :=\frac{1}{|\mathrm B|}
      \sum_{\x\in\mathrm B}
      \frac{\partial\log\tilde p(\x;\wparams)}
           {\partial\log\wparams_{n,c}},
    \qquad
    C_{\mathrm B}(n)
    :=\sum_{c\in\ch(n)}F_{\mathrm B}^{+}(n,c).
    \label{eq:app-pc-learning-flows}
\end{equation}
The inputs are held fixed, and derivatives treat untied
edge weights as independent coordinates before row
normalization. These derivatives are posterior expected
edge counts~\citep{peharz2020einsum}; $C_{\mathrm B}(n)$
is their total for node $n$.
Here $\tilde p$ denotes the raw circuit output, which
equals the model likelihood at normalized parameters.
The derivatives do not include an additional global
normalizer or a softmax parameterization.

Writing $T_{n,c}:=T_{\vparams,\wparams}(n,c)$, the
row-normalized form of \cref{eq:anemone} in this
standalone setting is
\begin{equation}
    \wparams^{\mathrm{new}}_{n,c}
    =\frac{\wparams_{n,c}
           +\alpha T_{n,c}F_{\mathrm B}^{+}(n,c)}
          {1+\alpha\sum_{d\in\ch(n)}
                     T_{n,d}F_{\mathrm B}^{+}(n,d)}.
    \label{eq:app-pc-learning-normalized}
\end{equation}
All quantities on the right are evaluated before the
update. The following choices establish the correspondence.

\paragraph{Full-batch EM.}
Let $\mathrm B=\mathcal D$ be the full training set and
take $T_{n,c}=1$. If $C_{\mathcal D}(n)>0$, then
\begin{equation}
    \lim_{\alpha\to\infty}
    \wparams^{\mathrm{new}}_{n,c}
    =\frac{F_{\mathcal D}^{+}(n,c)}{C_{\mathcal D}(n)}.
    \label{eq:app-pc-full-em}
\end{equation}
This is the EM update obtained by normalizing the expected
edge counts~\citep{peharz2020einsum}.
Thus, full-batch EM is recovered as a limit of the
additive update, rather than generally at a finite
learning rate.

\paragraph{Interpolated mini-batch EM.}
For $C_{\mathrm B}(n)>0$, choose
$T_{n,c}=1/C_{\mathrm B}(n)$ for every child $c$.
Substituting into \cref{eq:app-pc-learning-normalized}
gives
\begin{equation}
\begin{aligned}
    \wparams^{\mathrm{new}}_{n,c}
    &=\frac{\wparams_{n,c}
             +\alpha F_{\mathrm B}^{+}(n,c)/C_{\mathrm B}(n)}
            {1+\alpha}\\
    &=(1-\lambda)\wparams_{n,c}
      +\lambda\frac{F_{\mathrm B}^{+}(n,c)}{C_{\mathrm B}(n)},
    \qquad \lambda:=\frac{\alpha}{1+\alpha}.
\end{aligned}
    \label{eq:app-pc-minibatch-em}
\end{equation}
This recovers the mini-batch interpolation rule of
\citet{peharz2020einsum}, with mixing coefficient
$0\leq\lambda<1$. The endpoint $\lambda=1$ corresponds
to $\alpha\to\infty$.

\paragraph{Anemone.}
Let $\rho_n>0$ be the probability that a generative
traversal of the current circuit reaches node $n$.
Choosing $T_{n,c}=1/\rho_n$ gives
\begin{equation}
    \wparams^{\mathrm{new}}_{n,c}
    =\frac{\rho_n\wparams_{n,c}
           +\alpha F_{\mathrm B}^{+}(n,c)}
          {\rho_n+\alpha C_{\mathrm B}(n)},
    \label{eq:app-pc-anemone}
\end{equation}
which is Anemone's basic update
\citep[Eq.~(7)]{liu2025rethinkingprobabilisticcircuitparameter}.
Its effective step size depends on the node's usage in
the batch relative to its usage under the current model.

Rows with zero batch flow are left unchanged;
unreachable nodes are also left unchanged.
Nonnegative example weights, including $w(t)$, can be
included in the average in \cref{eq:app-pc-learning-flows},
yielding the corresponding weighted updates.
These correspondences concern the basic updates without
momentum or additional regularization.
For our conditional objective, the positive statistic
is replaced by $F_{\mathrm B}^{+}-F_{\mathrm B}^{-}$;
the dual-flow update addresses this signed case, with
its properties proved in Appendix~\ref{app:signed-flow-proof}.

\clearpage
\section{Proof of the Signed-Flow Update Properties}
\label{app:signed-flow-proof}
\begin{proof}[Proof of \cref{prop:signed-flow-update}]
Fix an active sum node $n$ and abbreviate
$\omega_c:=\omega_{n,c}$,
$a_c:=F_{\mathcal B}^{+}(n,c)$,
$b_c:=F_{\mathcal B}^{-}(n,c)$, and
$T_c:=T(n,c)$.
All sums below range over $c\in\ch(n)$.
The assumptions give $\omega_c>0$, $\sum_c\omega_c=1$,
finite $a_c,b_c\geq0$, and finite $T_c>0$.
All quantities are evaluated at the pre-update parameters
and held fixed as $\alpha$ varies.
The derivatives are taken with respect to directly parameterized,
untied edge weights before enforcing row normalization;
neural outputs and all other circuit inputs are held fixed.
Define the unnormalized updated weights and their row normalizer as
\begin{equation}
    u_c(\alpha)
    :=\omega_c
      \frac{1+\alpha T_ca_c/\omega_c}
           {1+\alpha T_cb_c/\omega_c},
    \qquad
    S(\alpha):=\sum_c u_c(\alpha).
    \label{eq:signed-proof-unnormalized}
\end{equation}
Then \cref{eq:signed-ratio-update} is precisely
$\omega_c^{\mathrm{new}}=u_c(\alpha)/S(\alpha)$.

\paragraph{(i) Positivity and normalization.}
For every finite $\alpha\geq0$, both terms
$1+\alpha T_ca_c/\omega_c$ and
$1+\alpha T_cb_c/\omega_c$ are finite and at least one.
Thus $u_c(\alpha)>0$ and $0<S(\alpha)<\infty$.
Consequently, $\omega_c^{\mathrm{new}}>0$ and
$\sum_c\omega_c^{\mathrm{new}}=1$.

\paragraph{(ii) Recovery of the standard update.}
Consider a smooth, decomposable standalone PC with normalized
input distributions and every sum row normalized.
Let $Z$ be its unconditional normalizer, with all variables
marginalized and no additional evidence factors. Let $\operatorname{TD}_{\wparams}(n)>0$ denote the probability that a generative traversal of the current
circuit visits sum node $n$. Then, we can write $\rho_n:=\operatorname{TD}_{\wparams}(n)>0$ for the probability that a generative traversal reaches $n$.
Under the independent-edge derivative convention,
\begin{equation}
    \frac{\partial\log Z}{\partial\log\omega_c}
    =\rho_n\omega_c.
    \label{eq:signed-proof-normalizer-flow}
\end{equation}
This identity is the normalizer derivative underlying the
standard circuit update
\citep{liu2025rethinkingprobabilisticcircuitparameter}.
To see it directly, evaluate the circuit with all variables
marginalized. Every node then has value one.
Reverse differentiation gives sensitivity $\rho_n$ at node $n$;
the derivative with respect to its edge weight $\omega_c$
is therefore $\rho_n$ times the child value, namely $\rho_n$.
Multiplication by $\omega_c/Z=\omega_c$ yields
\cref{eq:signed-proof-normalizer-flow}.
Although $Z=1$ at normalized parameters, independently
perturbing an edge need not preserve normalization, so this
derivative is generally nonzero.

For the weighted minibatch averages, let
$\bar w_{\mathcal B}:=|\mathcal B|^{-1}
\sum_{(t,\x_0,\x_t)\in\mathcal B}w(t)$.
In the standalone setting the unconditional normalizer is
the same for every example, giving
$b_c=\bar w_{\mathcal B}\rho_n\omega_c$.
Taking $T_c=1/\rho_n$ in
\cref{eq:signed-proof-unnormalized} therefore gives
\begin{equation}
    u_c(\alpha)
    =\frac{\omega_c+\alpha a_c/\rho_n}
           {1+\alpha\bar w_{\mathcal B}}.
    \label{eq:signed-proof-common-denominator}
\end{equation}
The denominator is shared across all children and cancels
under row normalization. Hence
\begin{equation}
    \omega_c^{\mathrm{new}}
    =\frac{\omega_c+\alpha a_c/\rho_n}
           {1+\alpha\sum_d a_d/\rho_n}
    =\frac{\rho_n\omega_c+\alpha a_c}
           {\rho_n+\alpha\sum_d a_d},
    \label{eq:signed-proof-standard-update}
\end{equation}
which is \cref{eq:anemone} with the positive flows $a_c$.
Here the standalone objective in \cref{eq:anemone} is
the average log circuit output, whose log-weight derivatives
are $a_c$.

\paragraph{(iii) Full-batch EM limit.}
Under the conditions in (ii), let $\mathcal B$ be the full
training set and assume $A:=\sum_d a_d>0$.
Dividing the numerator and denominator of
\cref{eq:signed-proof-standard-update} by $\alpha$ yields
\begin{equation}
    \lim_{\alpha\to\infty}\omega_c^{\mathrm{new}}
    =\lim_{\alpha\to\infty}
      \frac{\rho_n\omega_c/\alpha+a_c}
           {\rho_n/\alpha+A}
    =\frac{a_c}{A}.
\end{equation}
The $a_c$ are the corresponding weighted expected edge counts.
Normalizing them maximizes the sum-node EM surrogate
$\sum_c a_c\log\omega_c^{\mathrm{new}}$ subject to
nonnegative weights summing to one, and therefore gives
the full-batch EM update for the weighted objective.
For unit example weights, this is ordinary full-batch EM.
Zero counts may yield zero weights in this limit, which is
consistent with strict positivity holding only for finite steps.
If $A=0$, all $a_c$ vanish and
\cref{eq:signed-proof-standard-update} leaves the row unchanged;
the normalized-count formula is not used in that case.

\paragraph{(iv) First-order agreement.}
Let $G(n,c):=F_{\mathrm B}^{+}(n,c)-F_{\mathrm B}^{-}(n,c)$.
With all pre-update quantities fixed, define the
unnormalized update
$$
    r_{n,c}
    :=\omega_{n,c}
       \frac{1+\eta_{n,c}F_{\mathrm B}^{+}(n,c)}
            {1+\eta_{n,c}F_{\mathrm B}^{-}(n,c)},
$$
where $\eta_{n,c}=\alpha T(n,c)/\omega_{n,c}$ and
$\alpha\to0$. Then we can rewrite $r_{n,c}$ as $$r_{n,c} = \omega_{n,c}+\alpha T(n,c)G(n,c)+O(\alpha^2)$$ Since $\sum_d\omega_{n,d}=1$, the normalized
update satisfies
\begin{equation}
\begin{aligned}
    \omega_{n,c}^{\mathrm{new}}-\omega_{n,c}
    &=\frac{r_{n,c}-\omega_{n,c}\sum_d r_{n,d}}
            {\sum_d r_{n,d}}\\
    &=\frac{
        \omega_{n,c}+\alpha T(n,c)G(n,c)- \omega_{n,c}(1+\alpha\sum_dT(n,d)G(n,d)) + O(\alpha^2)
      }{
        1+\alpha\sum_dT(n,d)G(n,d)+O(\alpha^2)
      }\\
    &=\left[
        \alpha\left(T(n,c)G(n,c)
        -\omega_{n,c}\sum_dT(n,d)G(n,d)\right)
        +O(\alpha^2)
      \right]\\
    &\qquad\times
      \left[
        1-\alpha\sum_dT(n,d)G(n,d)+O(\alpha^2)
      \right]\\
    &=\alpha\left(T(n,c)G(n,c)
       -\omega_{n,c}\sum_dT(n,d)G(n,d)\right)
       +O(\alpha^2).
\end{aligned}
\end{equation}
All sums are over $d\in\ch(n)$.
This is also the first-order expansion of the normalized
signed-gradient update
$(\omega_{n,c}+\alpha T(n,c)G(n,c))/
 (1+\alpha\sum_dT(n,d)G(n,d))$,
establishing (iv).
\end{proof}

\end{document}